\documentclass[runningheads]{llncs}

\usepackage{eccv}

\usepackage{eccvabbrv}

\usepackage{graphicx}
\usepackage{booktabs}
\usepackage{multirow}
\usepackage{tabularx}
\usepackage{threeparttable}
\usepackage{svg}
\usepackage{amssymb}
\usepackage[accsupp]{axessibility}  

\usepackage{hyperref}

\usepackage{orcidlink}

\begin{document}

\title{SFPNet: Sparse Focal Point Network for Semantic Segmentation on General LiDAR Point Clouds} 

\titlerunning{SFPNet}

\author{Yanbo Wang\inst{1,2} \and
Wentao Zhao\inst{1,2} \and
Chuan Cao\inst{1,2} \and
Tianchen Deng\inst{1,2} \and
Jingchuan Wang\inst{1,2} \and
Weidong Chen\inst{1,2, \dag}} 
\renewcommand{\thefootnote}{}
\footnotetext[2]{$^\dag$Corresponding Author.}
\authorrunning{Y.~Wang et al.}

\institute{Institute of Medical Robotics and Department of Automation, Shanghai Jiao Tong University, China \\
\and{Key Laboratory of System Control and Information Processing, Ministry of Education, Shanghai 200240, China}}


\maketitle

\begin{abstract}

Although LiDAR semantic segmentation advances rapidly, state-of-the-art methods often incorporate specifically designed inductive bias derived from benchmarks originating from mechanical spinning LiDAR. This can limit model generalizability to other kinds of LiDAR technologies and make hyperparameter tuning more complex. To tackle these issues, we propose a generalized framework to accommodate various types of LiDAR prevalent in the market by replacing window-attention with our sparse focal point modulation. Our SFPNet is capable of extracting multi-level contexts and dynamically aggregating them using a gate mechanism. By implementing a channel-wise information query, features that incorporate both local and global contexts are encoded. We also introduce a novel large-scale hybrid-solid LiDAR semantic segmentation dataset for robotic applications. SFPNet demonstrates competitive performance on conventional benchmarks derived from mechanical spinning LiDAR, while achieving state-of-the-art results on benchmark derived from solid-state LiDAR. Additionally, it outperforms existing methods on our novel dataset sourced from hybrid-solid LiDAR. Code and dataset are available at \href{https://github.com/Cavendish518/SFPNet}{https://github.com/Cavendish518/SFPNet} and \href{https://www.semanticindustry.top}{https://www.semanticindustry.top}.
  
\end{abstract}

\section{Introduction}
\label{sec:intro}
\begin{figure}[tb]
  \centering
  \subfloat[Comparison of field of view.]{\includegraphics[height=2.2cm]{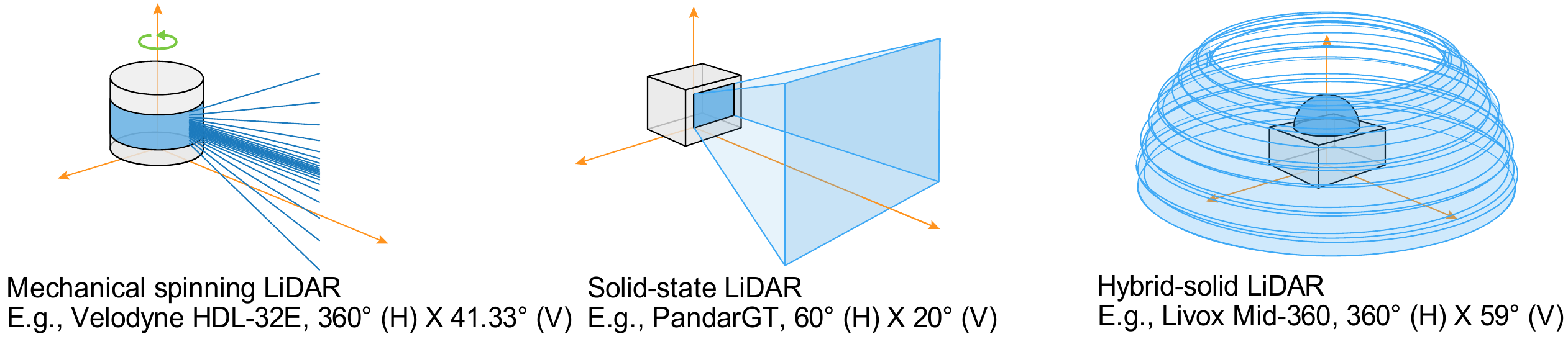}}
  \quad
  \subfloat[Comparison of cumulative point clouds.]{\includegraphics[height=1.7cm]{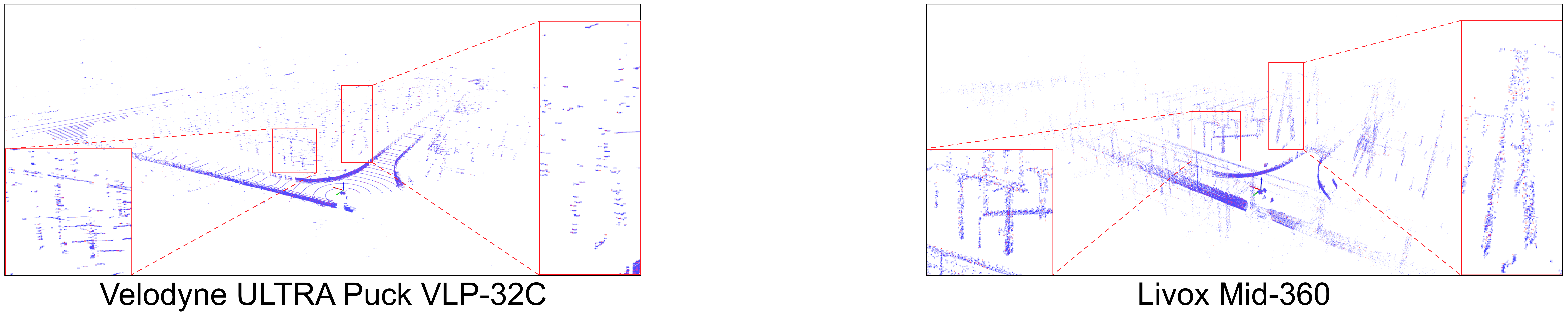}}
  
  \caption{Comparison of different types of LiDAR. \cref{fig:lidar} (a) compares three mainstream types of LiDAR technologies. Unlike camera, various types of LiDAR data have extremely different point distributions. Therefore, the generalizability of networks designed specifically for a particular LiDAR type is poor. \cref{fig:lidar} (b) contrasts the \textbf{cumulative} 1-second point clouds of Mid-360 (employed in our dataset) and commonly used VLP-32C. The non-repetitive scanning mode of Mid-360 covers a broader range of scenes, making it more suitable for industrial robots involving scene understanding tasks. Meanwhile, VLP-32C gathers more detailed road surface information.}
  \label{fig:lidar}
\end{figure}

Various types of 3D LiDAR sensors (as shown in \cref{fig:lidar}) have become popular choices in autonomous vehicles and robotics \cite{cadena2016past, chen2019suma++, gao2021we, xu2022fast} due to their accurate distance detection capabilities across diverse environments, including low-light conditions. The point clouds generated by LiDAR can accurately represent real-world scenes, facilitating direct 3D scene understanding through semantic segmentation. These advantages enable more effective support for subsequent tasks such as localization and planning compared to segmentation based on 2D images.

Despite the convenience afforded by LiDAR sensors, semantic segmentation based on LiDAR point clouds also encounters several challenges. These challenges primarily stem from characteristics inherent to LiDAR data. In general, the key features of all kinds of LiDAR data include sparsity, large scale, and non-uniform changes in point cloud density.

\begin{table}[tb]
  \caption{Comparison of representative frameworks. For each input format, the second example improves the performance by introducing inductive bias for mechanical spinning LiDAR. FPS\textsuperscript{\dag} and RS\textsuperscript{\ddag} are the abbreviations for farthest point sampling and random sampling, respectively.
  }
  \label{tab:headings}
  \centering
  \begin{threeparttable}
  \scalebox{0.8}{
  \begin{tabular}{@{}p{3.2cm}|l|l@{}}
    \hline
    Input Format & Example & Core inductive bias\\
    \cline{1-3}
    \multirow{2}*{Point clouds}  & PointNet++ \cite{qi2017pointnet++} & Neighbor consistency with FPS\textsuperscript{\dag}\\
    \cline{2-3}
    ~ & RandLA-Net \cite{hu2020randla} & Neighbor consistency with RS\textsuperscript{\ddag}\\
    \cline{1-3}
    \multirow{2}*{Range image}  & RangeNet++ \cite{milioto2019rangenet++} & Square locality\\
    \cline{2-3}
    ~ & RangeFormer \cite{kong2023rethinking} & Implicit square locality\\
    \cline{1-3}
    \multirow{2}*{\parbox{\linewidth}{Sparse voxel grid and points}}  & SPVNAS \cite{tang2020searching} & Cubic locality\\
    \cline{2-3}
    ~ & Cylinder3D \cite{zhu2021cylindrical} & Cylindrical and asymmetrical locality\\
    \cline{1-3}
    \multirow{2}*{\parbox{\linewidth}{Sparse voxel grid with point property}} & VoTr \cite{mao2021voxel} & Implicit cubic locality\\
    \cline{2-3}
    ~ & SphereFormer \cite{lai2023spherical} & Implicit radial and cubic locality\\
  \hline
  \end{tabular}
  }
  \end{threeparttable} 
\end{table}

Most of the pioneering work \cite{qi2017pointnet++,milioto2019rangenet++,zhu2021cylindrical,lai2023spherical} did not take into account all the characteristics of the LiDAR data, or the model capacity was insufficient, resulting in unsatisfactory performance. Cutting-edge works \cite{hu2020randla,kong2023rethinking,zhu2021cylindrical,lai2023spherical} adapt to the distribution of mechanical spinning LiDAR data through specially designed inductive bias as shown in \cref{tab:headings}.

However, there are various types of LiDAR with distinct characteristics as illustrated in \cref{fig:lidar}. According to the no free lunch theorem \cite{wolpert1997no}, state-of-the-art (SOTA) methods, which incorporate specifically designed inductive bias  (\eg, cylindrical partition \cite{zhu2021cylindrical} and radial window \cite{lai2023spherical} as shown in \cref{fig:compare}), risk constraining model generalizability and complicating hyperparameter tuning when applied to other types of LiDAR technologies.

Motivated by the analysis in \cref{fig:lidar}, we aim to propose a generalized framework capable of addressing the common characteristics of various types of LiDAR data prevalent in the market. Our goal is to ensure competitiveness on traditional benchmarks and demonstrate generality across other types of LiDAR data without introducing special inductive bias. 

Inspired by focal attention \cite{yang2021focal} and focal modulation \cite{yang2022focal}. We propose sparse focal point modulation (SFPM) by first extract features at different focal levels around each point. Then multi-level contexts are adaptively aggregated through a gate mechanism. Finally, a channel-wise information query is implemented to acquire the encoded features with both local and long-range information. Similar to window-attention \cite{lai2023spherical}, SFPM serves as a plugin module, thereby capable of seamlessly replacing window-attention in mainstream backbones.

To enable training and evaluation of LiDAR semantic segmentation based on hybrid-solid LiDAR, we build a new dataset, \textbf{S}e\textbf{M}antic \textbf{I}n\textbf{D}ustry (S.MID). We are the first to develop a hybrid-solid LiDAR semantic segmentation dataset based on the Livox Mid-360. To furnish robotic application scene for LiDAR semantic segmentation community, we used an industrial robot to collect a total of 38904 frames of LiDAR data in different substations. We annotated 25 categories under professional guidance and merged them into 14 classes for single-frame segmentation task.

\begin{sloppypar}
We evaluate our sparse focal point network (SFPNet) on two mechanical spinning LiDAR datasets, nuScenes \cite{caesar2020nuscenes} and SemanticKITTI \cite{behley2019semantickitti}. Our method achieves competitive results. We evaluate our SFPNet performance on solid-state LiDAR dataset through PandaSet \cite{xiao2021pandaset} and performance on hybrid-solid LiDAR dataset through our dataset S.MID. Our method achieves better results compared with the SOTA works. Experimental results show that our frameworks have strong generalization capability and interpretability.    
\end{sloppypar}

We summarize our contributions as below:
\begin{itemize}
    \item{We introduce SFPNet for feature encoding of sparse point clouds obtained from various types of LiDAR sensors. SFPNet effectively avoids introducing special inductive bias while ensuring expansive receptive fields. Additionally, it offers enhanced interpretability for semantic segmentation tasks.} 
    \item{In our sparse focal point modulation, the introduced multi-level feature extraction and gated aggregation can adaptively learn local and global features from various LiDAR point cloud data with different distribution patterns and non-uniform density variations.}
    \item{A novel dataset for LiDAR semantic segmentation has been collected. S.MID is built with a novel hybrid-solid LiDAR in the substations. It fills the gap of public dataset in industrial outdoor scenes for robotic application.}
    \item{Our proposed method achieves the cutting-edge results on both nuScenes and SemanticKITTI, which are based on mechanical spinning LiDAR. More importantly, our new framework demonstrates its ability to generalize across different LiDAR technologies, as evidenced by its leading performance on PandaSet (solid-state LiDAR) and S.MID (hybrid-solid LiDAR).}
\end{itemize}

\section{Related Works}
From advancements in 3D point cloud recognition to the development of LiDAR-based semantic segmentation, a variety of interesting techniques have been introduced. In accordance with the input format, methods are typically classified into point-based, projection-based, voxel-based, and multi-modality-based approaches. Previous high-performance methods usually have designed networks with tailored inductive bias to effectively address the characteristics of LiDAR point cloud data, including its large scale, low data volume, overall sparsity, and non-uniform density variations. In this section, we aim to summarize previous works from a novel inductive bias view.
\subsection{Explicit Locality Assumption Methods}
The explicit locality assumptions of mainstream frameworks are usually derived from the inherent properties of K-nearest neighbor (KNN), 2D CNN, 3D CNN and submanifold sparse convolutional networks (SSCN) \cite{graham20183d}. 

\subsubsection{Explicit 2D Locality.}
Explicit 2D locality assumption methods usually employ a 2D backbone to extract features from a range view \cite{milioto2019rangenet++,behley2019semantickitti,wu2018squeezeseg,wu2019squeezesegv2,xu2020squeezesegv3,aksoy2020salsanet,razani2021lite}, a bird-eye view (BEV) \cite{zhang2020polarnet}, or a combination of three planes \cite{puy2023using} projected from LiDAR point clouds. In addition to the information loss caused during the projection process, the inductive bias of traditional 2D CNN also suffers from a certain degree of failure. Not only does the 2D CNN process irrelevant points in faraway locations, rendering locality invalid, but the deformation produced in the projection space also does not satisfy the spatial invariance assumption. Although the structure relying on explicit 2D locality is efficient, the above problems make the model capacity of this type of method easily reach the upper limit.

\subsubsection{Explicit 3D Locality.} 
Explicit 3D locality assumption methods usually adopt a PointNet++ \cite{qi2017pointnet++} like network or SSCN \cite{graham20183d} to extract features. The former kinds of methods \cite{hu2020randla,thomas2019kpconv,qian2022pointnext,yan2020pointasnl,tatarchenko2018tangent} rely on neighbor consistency in 3D point clouds to extract and aggregate multi-scale features. The latter kinds of methods \cite{tang2020searching,cheng20212,choy20194d,jiang2021guided,cohen2018spherical,liu2022less,yan2021sparse} depend on the inductive bias from 3D CNN or SSCN. However, non-uniform density variations of LiDAR point clouds limit the effectiveness of SSCN. Further improvements have been made by modifying the output region \cite{chen2022focal} or enlarging the receptive field \cite{lu2023link,chen2023largekernel3d}. Cylinder3D \cite{zhu2021cylindrical} proposes a cylindrical partition by investigating point distribution obtained from mechanical spinning LiDAR to make the data more consistent with the locality assumption of SSCN. In addition, the introduced asymmetrical convolution design \cite{zhu2021cylindrical} is more consistent with the density changes of point clouds in each direction. We have noticed that the key to the success of these improved methods enable SSCN to operate in a way that is more suitable for data distribution to obtain favorable properties such as better 3D locality assumptions.

\subsection{Implicit Locality Assumption Methods}
Transformer \cite{vaswani2017attention} has shown strong capability in modeling long-range dependency with the self-attention mechanism. However, it requires large dataset and longer training schedules to learn the implicit locality. Leveraging the attributes of mechanical spinning LiDAR, various architectures, tokenization strategies, window forms and positional encoding for transformer have been proposed.

\subsubsection{Implicit 2D Locality.}
Recently, RangeViT \cite{ando2023rangevit} and RangeFormer \cite{kong2023rethinking} have been proposed to overcome the problem of insufficient model capacity of the explicit 2D locality assumption methods through the power of the transformer architecture. However, the performance still lags behind the latest 3D methods, requiring reliance on pre-processing and post-processing to achieve SOTA results.

\subsubsection{Implicit 3D Locality.} 
Several pioneering transformer-based works \cite{mao2021voxel, wu2022point, fan2022embracing, lai2022stratified} have been proposed to solve 3D point cloud perception. These works take 3D point cloud sequences or sparse voxel with point properties as input and learn 3D locality implicitly. These methods are more suitable for indoor datasets than LiDAR datasets. SphereFormer \cite{lai2023spherical} introduces locality in the radial direction by calculating self-attention within the radial window to improve the perception of distant LiDAR point clouds. We have noticed that the key to the success of these methods is the introduction of long-range dependency while implicitly learning locality through specially designed positional encoding.

\subsection{Multi-modality Assumption Methods} \label{sec:2.3}
Fundamental assumptions in a single modality have drawbacks as mentioned above. Introducing additional modalities is regarded as a solution to address the limitation in model capacity arising from the assumption of a singular modality. RPVNet \cite{xu2021rpvnet} takes advantage of multi-view fusion. This work leverages an inductive bias towards arbitrary relations between views, facilitating the flow of information across different views at the feature level. Moreover, \cite{yan20222dpass,liu2023uniseg,genova20212d3dnet,zhuang2021perception} introduce additional image information, which inherits the inherent inductive bias of the fundamental backbones. Although our work is dedicated to feature encoding of general LiDAR point clouds, our approach still achieved superior performance than some of these methods.

Through summarizing the structure designed by cutting-edge frameworks, we propose our network architecture under explicit locality assumption with adaptive hierarchical contextualization aggregation to capture long-range dependence and global information. We take sparse voxel with point properties as our only input. Please note that our work focuses on the representational capabilities of the network design itself, rather than special data augmentation \cite{xu2021rpvnet,xiao2022polarmix,xu2023frnet, kong2023rethinking,lai2023spherical}, training skills \cite{kong2023lasermix}, post-processing \cite{milioto2019rangenet++, ando2023rangevit} and distillation \cite{hou2022point,liu2022less,yan20222dpass}. Therefore, related powerful training techniques are not summarized here and have not been used in the experiments.

\section{Methods}
In this section, we first summarize the sparse submanifold convolution and self-attention mechanism employed in mainstream backbones and compare them with the sparse focal point modulation introduced in our research, as outlined in \cref{rethink}. Subsequently, we provide details for the implementation of sparse focal point modulation in \cref{sfpm}. Then, we give our overall framework and objective function in the supplementary materials.

\subsection{Rethinking Submanifold Sparse Convolution and Window-Attention} \label{rethink}
In general, considering a sorted LiDAR feature sequence $X \in \mathbb{R}^{N \times C_{in}}$ (feature in sparse tensor or feature inherited from original point sequence) as input, the output feature $y_{i} \in \mathbb{R}^{C_{out}}$ is encoded from token (or voxel) $x_{i} \in \mathbb{R}^{C_{in}}$ through neighborhood interaction $\xi$  and contextual aggregation $\kappa$.

\subsubsection{Submanifold Sparse Convolution.}
\begin{sloppypar}
The submanifold sparse convolution without pooling operation can be formulated as:    
\end{sloppypar}
\begin{equation}
y_{i} = \xi_{SubMconv} (i,X ),
\label{eq:SubMconv}
\end{equation}
where neighbourhood features at location $i$ are captured with well preserved sparsity \cite{graham2017submanifold} through efficient interaction $\xi_{SubMconv}$.

\subsubsection{Window-Attention.}
The window-attention \cite{liu2021swin} can be formulated as:
\begin{equation}
y_{i} = \kappa_{attn} (x_{i}, \xi_{attn} (x_{i},X)),
\label{eq:attn}
\end{equation}
where the computation of attention scores between the query and key via the interaction $\xi_{attn}$ occurs before the aggregation $\kappa_{attn}$ within the window.

\subsubsection{Sparse Focal Point Modulation.}
\begin{sloppypar}
Followed by focal modulation \cite{yang2022focal} paradigm, sparse focal point modulation (SFPM) can be formulated as:
\end{sloppypar}

\begin{equation}
y_{i} = \xi_{focal} (x_{i}, \kappa_{focal} (i,X)),
\label{eq:focal}
\end{equation}
where the interaction operation $\xi_{focal}$ performs the query task subsequent to aggregated contexts of the focal neighbors using $\kappa_{focal}$ at each location $i$.

\begin{figure}[tb]
  \centering
  \includegraphics[height=4.5cm]{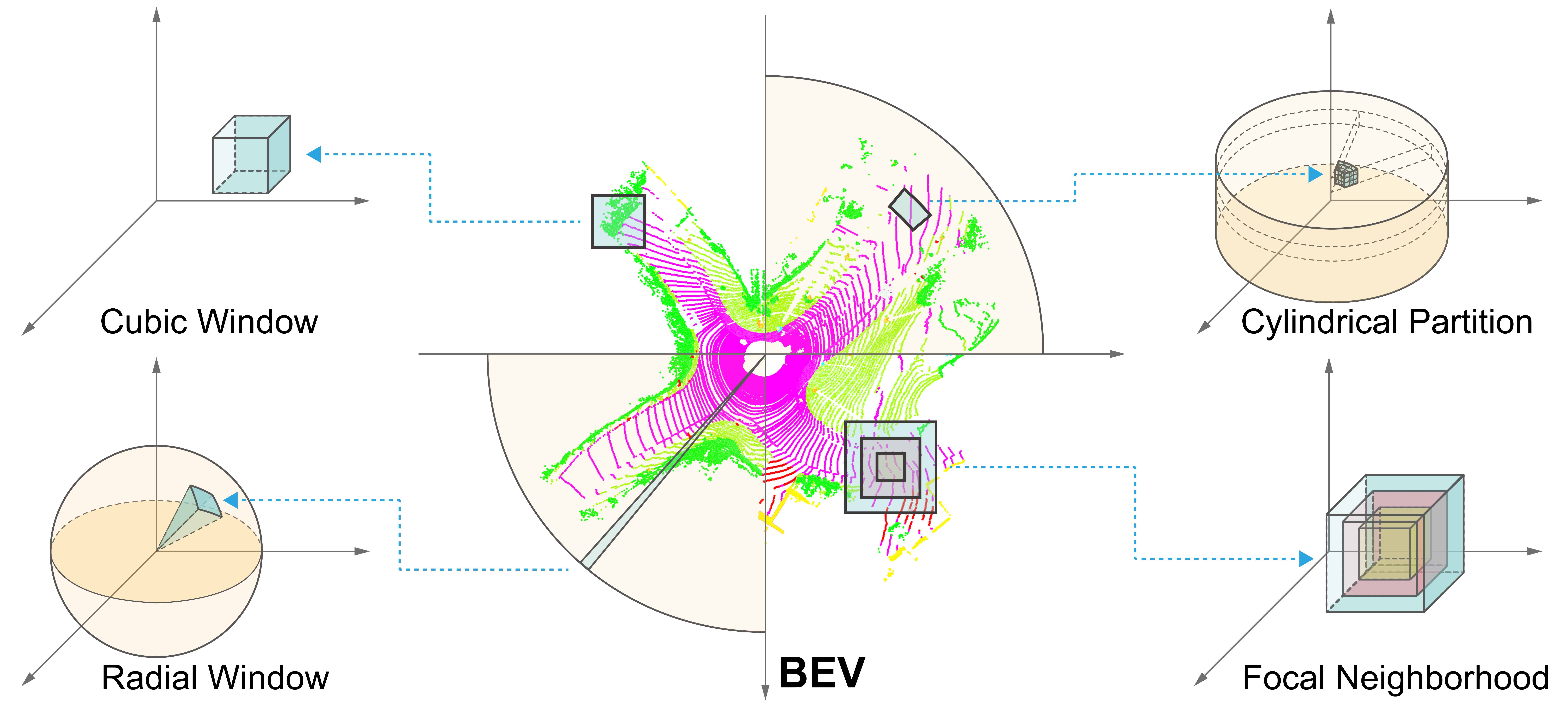}
  \caption{Heuristic comparison with mainstream design. Based on point distribution of mechanical spinning LiDAR, Cylindrical partition \cite{zhu2021cylindrical} and radial window \cite{lai2023spherical} are proposed to extract the features of distant points. Focal neighborhood adapts to this problem by aggregating multi-level contexts. Since no special inductive bias is introduced, it can be elegantly applied to various kinds of LiDAR as shown in \cref{fig:lidar}.}
  \label{fig:compare}
\end{figure}

\subsubsection{Comparison with Mainstream Design.} 
As an extension of \cite{yang2022focal}, we believe that SFPM combines the advantages of submanifold sparse convolution and window-attention by comparing \cref{eq:SubMconv,eq:attn,eq:focal}.

\begin{itemize}
    \item{ \emph{Explicit locality with contextual learning.} $\kappa_{focal}$ extracts the contexts from location $i$. $\xi_{focal}$ preserves channel information for each token $x_{i}$. Therefore, SFPM simultaneously possesses spatial-specific and channel-specific properties while exhibiting decoupled feature granularity.} 
    \item{\emph{Translation invariance.} SFPM is invariant to translation of sorted input LiDAR feature sequence $X$, since operation $\xi_{focal}$ and $\kappa_{focal}$ are always centered at location $i$. This also eliminates the need for positional encoding. }
\end{itemize}

A heuristic comparison with mainstream design between our methods with Cylinder3D \cite{zhu2021cylindrical} and SphereFormer \cite{lai2023spherical} can be found in \cref{fig:compare}. With the above advantages, it does not rely on special point distribution assumption, and has a large receptive field.

\subsection{Sparse Focal Point Modulation} \label{sfpm}
\begin{figure}[tb]
  \centering
  \includegraphics[width=\linewidth]{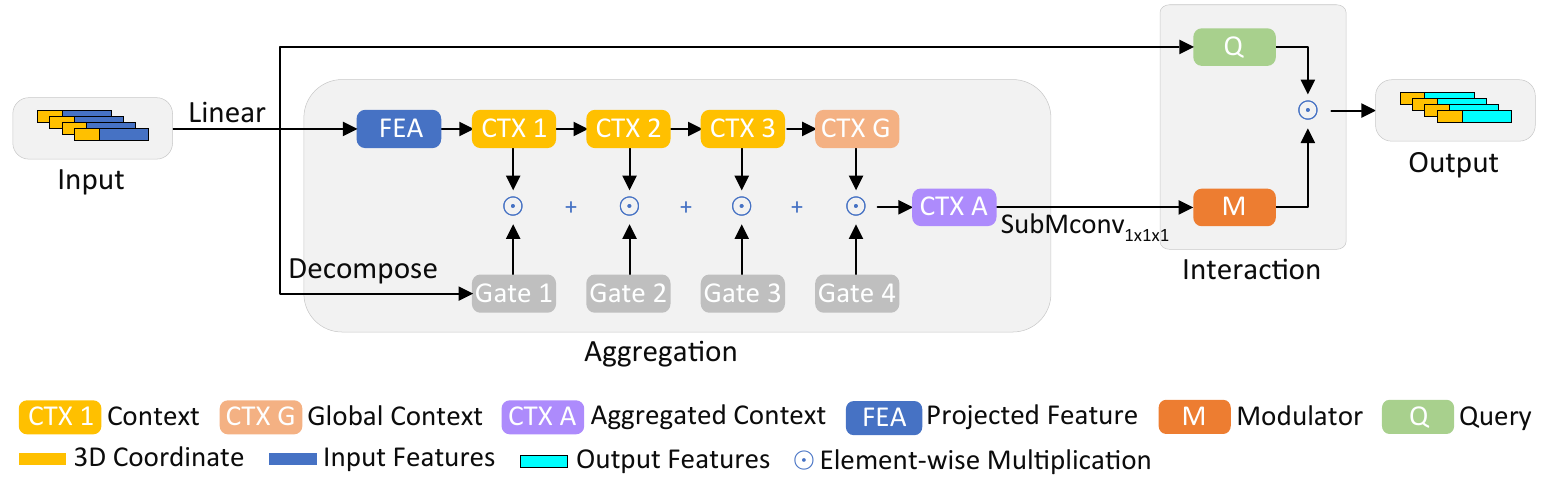}
  \caption{Illustration of sparse focal point modulation.}
  \label{fig:sfpm}
\end{figure}

\cref{fig:sfpm} shows the structure of SPFM. Based on \cref{eq:focal}, $\kappa_{focal}$ is designed through \textbf{multi-level context extraction} and \textbf{adaptive feature aggregation}, and $\xi_{focal}$ is designed into \textbf{channel-wise information query}.
\subsubsection{Multi-level Context Extraction.}
Previous works have proved that both local features and long-range contexts are important for LiDAR segmentation \cite{lai2023spherical, lu2023link, chen2023largekernel3d}. Therefore, we hope to obtain features of different levels in the first step to acquire hierarchical information.

Given a sorted LiDAR feature sequence $X \in \mathbb{R}^{N \times C_{in}}$, we first employ a linear layer $S^{0} = f_{X}^{S}(X) \triangleq MLP (X)$ to project it into a new feature space while keeping the same channel numbers $C_{in}$. Then the sparse multi-level contexts can be obtained through a sequence of 3D submanifold sparse convolution with $C_{in} = C_{out}$ and a kernel size of $k^{l}$ at level $l$. 

At focal level $l\in \{1,...,L\}$, the output $S^{l}$ is derived by:
\begin{equation}
    S^{l} = f_{l-1}^{l}(S^{l-1}) \triangleq LN(GeLU(SubMconv^{l}_{3d}(S^{l-1}))),
\label{eq:context}
\end{equation}
where LN represents for  layer normalization \cite{ba2016layer}, and GeLU is Gaussian error linear unit \cite{hendrycks2016gaussian}. The kernel size at each focal level is increasing through $k^{l} = k^{l-1} + 2$. The effective receptive field at each level is $RF^{l} = 1 +  {\textstyle \sum_{i=1}^{l}} (k^{l}-1)$, which leads to long-range dependence learning. 

Finally, global average pooling $S^{l+1} =Avgpool(S^{l})$ is performed on the $L$-$\ts{th}$ level features to obtain global context. 

\subsubsection{Adaptive Feature Aggregation.}
Through the above steps,  multi-level contexts $\{S^{1},...,S^{L},S^{L+1}\}$ have been extracted. However, not all the contexts are equally important. Even point clouds from the same type of LiDAR belonging to the same type of objects will produce different multi-level features due to various point cloud distributions. Therefore, an adaptive feature aggregation is achieved through gated mechanism. Following the gated aggregation in \cite{yang2022focal}, the spatial-aware gating weights for each level are calculated through $G = f_{X}^{G}(x) \triangleq MLP (X)$ with $L+1$ channels. And the cross channel result $F^{out}$ after gate aggregation is obtained through: 
\begin{align}
    S^{out} &=\sum_{l=1}^{L+1} G^{l}\odot S^{l}, \label{eq: gate} \\
    F^{out} &= h(S^{out}) \triangleq SubMconv^{1\times 1\times 1}_{3d}(S^{out}), \label{eq: int}
\end{align}
where $\odot$ is element-wise multiplication and $h( \cdot )$ represents the cross channel aggregation. 

The whole aggregation step $\kappa_{focal}$ can adaptively learn hard tokens through multi-level contexts, and can also avoid introducing too much invalid information to easy tokens. This also enables SFPM to accommodate various distribution patterns of point clouds without special inductive bias. Model interpretation through visualization experiments can be found in \cref{exp ana}.

\subsubsection{Channel-wise Information Query.}
Following \cite{yang2021focal, yang2022focal, vaswani2017attention,liu2021swin}, $\xi_{focal}$ is simply implemented through a query projection $q(x_{i}) \triangleq MLP (X)$ with the same channel number $C_{in} = C_{out}$. Derived from \cref{eq:focal,eq:context,eq: gate,eq: int}, the features from SFPM is encoded via:
\begin{equation}
    y_{i} = q(x_{i})\odot h(\sum_{l=1}^{L+1} g_{i}^{l} \cdot s_{i}^{l}),
\end{equation}
where $g_{i}^{l}$ and $s_{i}^{l}$ are the gate weights and sparse contexts at location $i$ and level $l$, respectively. Through this lightweight element-wise multiplication, channel-wise information is well-preserved.

\section{Experiments}
In this section, we first introduce our experiment setups in \cref{exp setup}. Then we show segmentation results across different datasets in \cref{exp results}. We further analyze the network design and interpretability in \cref{exp ana}. Finally, ablation study is shown in \cref{exp abl}. 
\subsection{Experiment Setups} \label{exp setup}
\subsubsection{Datasets.}
\begin{figure}[tb]
\centering
\includegraphics[width=\textwidth]{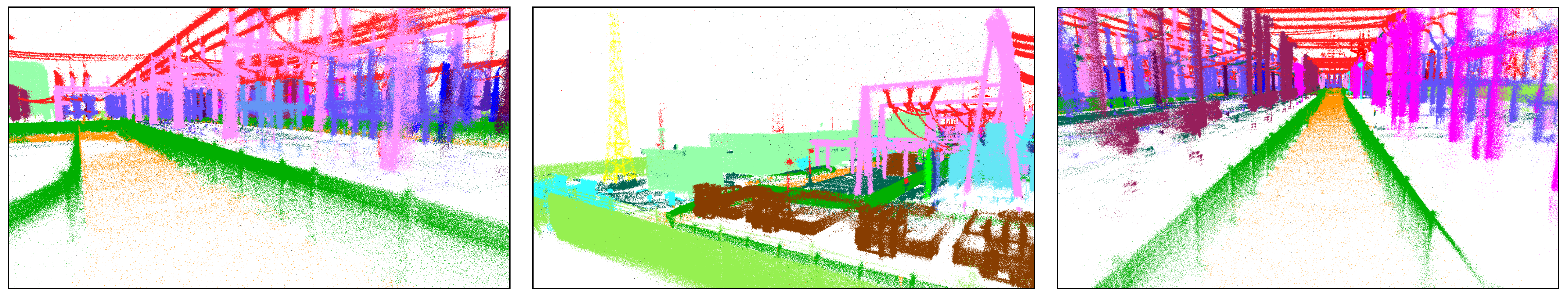}
\caption{Labeled cumulative point clouds in our novel dataset \textbf{ S.MID}. We provide dense semantic annotations for each frame.}
\label{fig:dataset}
\end{figure}

\begin{itemize}
    \begin{sloppypar}
        \item{\emph{Mechanical Spinning LiDAR.} Two large-scale driving-scene benchmarks, nuScenes (Velodyne-HDL-32E, 16 classes) \cite{caesar2020nuscenes} and SemanticKITTI \cite{behley2019semantickitti} (Velodyne-HDL-64E, 19 classes) are employed to verify the performance for point clouds obtained from traditional mechanical spinning LiDAR.} 
    \end{sloppypar}
    \item{\emph{Solid-State LiDAR.} We extract labeled solid-state LiDAR data from PandaSet (PandarGT) \cite{xiao2021pandaset}. We merge and select 13 classes for evaluation. The data split is in an 4:1 ratio for training and validation.}
    \item{\emph{Hybrid-Solid LiDAR.} We develop a novel dataset S.MID. We hope to setup a large-scale robotic application benchmark for LiDAR semantic segmentation task. We collect a total of 38904 frames of LiDAR data in different substations through an industrial robot equipped with Livox Mid-360, a novel hybrid-solid LiDAR. We carefully split the dataset as follows. 13,101 frames are collected from one complete substation for training. Validation and test sets are sourced from different substations, comprising 5,000 frames and 20,803 frames, respectively. 25 categories are annotated under professional guidance as shown in \cref{fig:dataset}. 14 classes are remained for single-frame segmentation task after merging classes with collective name and ignore classes with very few points. More details (sensors, scenes, annotation process, label distributions, \etc) about datasets can be found in the supplementary materials.}
\end{itemize}

\subsubsection{Implementation Details.}
We utilize two GeForce RTX 3090 GPUs for training, with the exception of SemanticKITTI, where four GPUs are employed. We train our models for 70 epochs using a batch size of 8, employing the AdamW optimizer \cite{loshchilov2017decoupled} with a learning rate of 0.0008 and a polynomial learning rate scheduler.

\subsection{Segmentation Results} \label{exp results}
As mentioned in \cref{sec:2.3}, we focus on the representational capabilities of the network design itself and have not employed powerful training techniques in the experiments. Additionally, we select open source SOTA methods \cite{zhu2021cylindrical,lai2023spherical} as our main comparison methods, since these methods play an important role as backbones or one of the modalities in cutting-edge works. 

\subsubsection{Mechanical Spinning LiDAR.}
\begin{sloppypar}
The semantic segmentation results for the nuScenes and SemanticKITTI validation and test sets are displayed in \cref{tab:valall}, \cref{tab:nuScenes}, and supplementary materials. The results demonstrate that our approach can achieve competitive performance even without inductive bias specifically tailored for mechanical spinning LiDAR. We attribute this achievement to the robust adaptability of our newly designed SFPM module. As shown in \cref{tab:nuScenes}, our method outperforms all existing LiDAR-based methods on the nuScenes validation set. Compared to the methods incorporating explicit 2D locality assumptions \cite{milioto2019rangenet++, zhang2020polarnet, cortinhal2020salsanext,Liong2020AMVNetAM}, ours yields a substantial performance improvement, ranging from 4.0\% \~{} 14.6\% in terms of mIoU.  Moreover, thanks to multi-level context aggregation, our method surpasses the models relying on explicit 3D locality assumptions \cite{cheng20212, park2023pcscnet,zhao2022svaseg,zhu2021cylindrical} by 4.0\% \~{} 17.9\% mIoU. Our method also achieves 0.6\% \~{} 4.9\% mIoU performance gain compared to methods with implicit 2D/3D locality assumptions  \cite{ando2023rangevit,kong2023rethinking,lai2023spherical}. It is notable that our LiDAR-based method even outperforms several approaches utilizing additional 2D information \cite{zhuang2021perception,genova20212d3dnet,yan20222dpass}.
\end{sloppypar}

\begin{table}[tb]
  \caption{Comparison with SOTA backbone type of works on four datasets. Note that all results are obtained from the literature\textsuperscript{\dag} or from open source codes\textsuperscript{\ddag} with carefully chosen parameters. No powerful training techniques are employed during our model training.}
  \label{tab:valall}
  \centering
  \scalebox{0.7}{
  \begin{tabular}{l|c|c|c|c|c|c|c}
    \hline
    \rule{0pt}{10pt}
    \multirow{2}{*}{\makebox[2.6cm][c]{Methods}} 
    & \multicolumn{2}{c|}{nuScenes\textsuperscript{\dag}} & \multicolumn{2}{c|}{SemanticKITTI\textsuperscript{\dag}} & {PandaSet\textsuperscript{\ddag}} & \multicolumn{2}{c}{S.MID\textsuperscript{\ddag}} \\
    \cline{2-8}
    & \makebox[1.3cm][c]{Val} & \makebox[1.3cm][c]{Test} & \makebox[1.3cm][c]{Val} & \makebox[1.3cm][c]{Test} & \makebox[1.3cm][c]{Val}  & \makebox[1.3cm][c]{Val} & \makebox[1.3cm][c]{Test} \\
    \hline
    \hline
    Cylinder3D \cite{zhu2021cylindrical} & 76.1 & 77.2 & 65.9 & 68.9 & 55.0 & 68.8 & 68.1 \\
    \hline
    SphereFormer \cite{lai2023spherical} & 79.5 & 81.9 & 69.0 & 74.8 & 63.5 & 67.8 & 68.3 \\
    \hline
    \hline
    Ours & \textbf{80.1}  & 80.2 & \textbf{69.2} & 70.3 & \textbf{64.0} & \textbf{71.9} & \textbf{70.9} \\
    \hline
  \end{tabular}
  }
\end{table}

\begin{table*}[t]
    \caption{Results of our proposed method and SOTA LiDAR Segmentation methods on nuScenes val set. Note that all results are obtained from the literature. All of the LiDAR-based methods published before March 7, 2024 have been listed. L and C represent for LiDAR and camera, respectively. Top 3 for each class are marked in \textcolor{blue}{blue}.}
    \label{tab:nuScenes}
    \centering
    \resizebox{\linewidth}{!}{
    \begin{tabular}{c|c|c|c|c|c|c|c|c|c|c|c|c|c|c|c|c|c|c}
        \hline
        \textbf{Methods (year)} & \textbf{Input} &\textbf{mIoU} & \rotatebox{90}{barrier} &  \rotatebox{90}{bicycle} & \rotatebox{90}{bus} & \rotatebox{90}{car} & \rotatebox{90}{construction} & \rotatebox{90}{motorcycle} & \rotatebox{90}{pedestrian} & \rotatebox{90}{traffic cone} & \rotatebox{90}{trailer} & \rotatebox{90}{truck} & \rotatebox{90}{driveable} & \rotatebox{90}{other flat} &
        \rotatebox{90}{sidewalk} & \rotatebox{90}{terrain} & \rotatebox{90}{manmade} & \rotatebox{90}{vegetation} \\
        \hline
        \hline
        (AF)$^{2}$-S3Net \cite{cheng20212} ('21) & L & 62.2 & 60.3 & 12.6 & 82.3 & 80.0 & 20.1 & 62.0 & 59.0 & 49.0 & 42.2 & 67.4 & 94.2 & 68.0 & 64.1 & 68.6 & 82.9 & 82.4
        \\ \hline
        RangeNet53++ \cite{milioto2019rangenet++} ('19) & L & 65.5 & 66.0 & 21.3 & 77.2 & 80.9 & 30.2 & 66.8 & 69.6 & 52.1 & 54.2 & 72.3 & 94.1 & 66.6 & 63.5 & 70.1 & 83.1 & 79.8
        \\ \hline
        PolarNet \cite{zhang2020polarnet} ('22) & L & 71.0 & 74.7 & 28.2 & 85.3 & 90.9 & 35.1 & 77.5 & 71.3 & 58.8 & 57.4 & 76.1 & 96.5 & 71.1 & 74.7 & 74.0 & 87.3 & 85.7
        \\ \hline
        PCSCNet \cite{park2023pcscnet} ('22) & L & 72.0 & 73.3 & 42.2 & 87.8 & 86.1 & 44.9 & 82.2 & 76.1 & 62.9 & 49.3 & 77.3 & 95.2 & 66.9 & 69.5 & 72.3 & 83.7 & 82.5        
        \\ \hline
        SalsaNext \cite{cortinhal2020salsanext} ('20) & L & 72.2 & 74.8 & 34.1 & 85.9 & 88.4 & 42.2 & 72.4 & 72.2 & 63.1 & 61.3 & 76.5 & 96.0 & 70.8 & 71.2 & 71.5 & 86.7 & 84.4
        \\ \hline
        SVASeg \cite{zhao2022svaseg} ('22) & L & 74.7 & 73.1 & 44.5 & 88.4 & 86.6 & 48.2 & 80.5 & 77.7 & 65.6 & 57.5 & 82.1 & 96.5 & 70.5 & 74.7 & 74.6 & 87.3 & 86.9 
        \\ \hline
        RangeViT-CS \cite{ando2023rangevit} ('23) & L & 75.2 & 75.5 & 40.7 & 88.3 & 90.1 & 49.3 & 79.3 & 77.2 & 66.3 & 65.2 & 80.0 & 96.4 & 71.4 & 73.8 & 73.8 & 89.9 & 87.2
        \\ \hline
        AMVNet \cite{Liong2020AMVNetAM} ('20) & L & 76.1 & \textcolor{blue}{79.8} & 32.4 & 82.2 & 86.4 & \textcolor{blue}{62.5} & 81.9 & 75.3 & \textcolor{blue}{72.3} & \textcolor{blue}{83.5} & 65.1 & \textcolor{blue}{97.4} & 67.0 & \textcolor{blue}{78.8} & 74.6 & 90.8 & 87.9
        \\ \hline
        Cylinder3D \cite{zhu2021cylindrical} ('21) & L & 76.1 & 76.4 & 40.3 & 91.2 & \textcolor{blue}{93.8} & 51.3 & 78.0 & 78.9 & 64.9 & 62.1 & 84.4 & 96.8 & 71.6 & \textcolor{blue}{76.4} & 75.4 & 90.5 & 87.4
        \\ \hline
        PVKD \cite{hou2022point} ('22) & L & 76.0 & 76.2 & 40.0 & 90.2 & \textcolor{blue}{94.0} & 50.9 & 77.4 & 78.8 & 64.7 & 62.0 & 84.1 & 96.6 & 71.4 & \textcolor{blue}{76.4} & \textcolor{blue}{76.3} & 90.3 & 86.9
        \\ \hline
        PMF \cite{zhuang2021perception} ('21) & L+C & 76.9 & 74.1 & 46.6 & 89.8 & 92.7 & 57.0 & 77.7 & 80.9 & \textcolor{blue}{70.9} & 64.6 & 82.9 & 95.5 & 73.3 & 73.6 & 74.8 & 89.4 & 87.7
        \\ \hline
        RPVNet \cite{xu2021rpvnet} ('21) & L & 77.6 & 78.2 & 43.4 & 92.7 & 93.2 & 49.0 & \textcolor{blue}{85.7} & 80.5 & 66.0 & 66.9 & 84.0 & 96.9 & 73.5 & 75.9 & \textcolor{blue}{76.0} & 90.6 & 88.9
        \\ \hline
        RangeFormer \cite{kong2023rethinking} ('23) & L & 78.1 & 78.0 & 45.2 & 94.0 & 92.9 & 58.7 & 83.9 & 77.9 & 69.1 & 63.7 & 85.6 & 96.7 & \textcolor{blue}{74.5} & 75.1 & 75.3 & 89.1 & 87.5        
        \\ \hline
        2D3DNet \cite{genova20212d3dnet} ('21)& L+C & 79.0 & 78.3 & \textcolor{blue}{55.1} & \textcolor{blue}{95.4} & 87.7 & 59.4 & 79.3 & 80.7 & \textcolor{blue}{70.2} & 68.2 & \textcolor{blue}{86.6} & 96.1 & \textcolor{blue}{74.9} & 75.7 & 75.1 & \textcolor{blue}{91.4} & \textcolor{blue}{89.9}
        \\ \hline
        2DPASS \cite{yan20222dpass} ('22) & L(C) & 79.4 & \textcolor{blue}{78.8} & \textcolor{blue}{49.6} & \textcolor{blue}{95.6} & 93.6 & \textcolor{blue}{60.0} & 84.1 & \textcolor{blue}{82.2} & 67.5 & 72.6 & \textcolor{blue}{88.1} & 96.8 & 72.8 & 76.2 & \textcolor{blue}{76.5} & 89.4 & 87.2
        \\ \hline
        SphereFormer \cite{lai2023spherical} ('23) & L & 79.5 & 78.7 & 46.7 & 95.2 & \textcolor{blue}{93.7} & 54.0 & \textcolor{blue}{88.9} & \textcolor{blue}{81.1} & 68.0 & \textcolor{blue}{74.2} & 86.2 & \textcolor{blue}{97.2} & 74.3 & 76.3 & 75.8 & \textcolor{blue}{91.4} & \textcolor{blue}{89.7}
        \\ \hline \hline
        Ours & L & \textbf{80.1} & \textcolor{blue}{78.8} & \textcolor{blue}{49.7} & \textcolor{blue}{95.3} & 93.5 & \textcolor{blue}{63.1} & \textcolor{blue}{86.4} & \textcolor{blue}{82.9} & 68.6 & \textcolor{blue}{72.8} & \textcolor{blue}{86.7} & \textcolor{blue}{97.0} & \textcolor{blue}{74.7} & 76.0 & 75.3 & \textcolor{blue}{91.2} & \textcolor{blue}{89.5}
        \\ \hline
    \end{tabular}
    }
\end{table*}

\subsubsection{Solid-State LiDAR.}
The results on PandaSet val set are shown in \cref{tab:Panda}. In comparison to mechanical spinning LiDAR, solid-state LiDAR exhibits a smaller horizontal field of view, greater detection range, and much finer vertical resolution. Our model surpasses the Cylinder3D \cite{zhu2021cylindrical}, achieving a notable margin of 9.0\% mIoU. Even when compared to the SphereFormer \cite{lai2023spherical} which is proposed specifically for enhancing long-range dependencies fit for greater detection range, our approach still maintains a lead of 0.5\% mIoU.

\begin{table*}[t]
    \caption{Results of our proposed method and SOTA LiDAR Segmentation methods on PandaSet validation set. Note that all results are obtained from open source code with carefully chosen parameters.}
    \label{tab:Panda}
    \centering
    \resizebox{0.65\linewidth}{!}{
    \begin{tabular}{c|c|c|c|c|c|c|c|c|c|c|c|c|c|c}
    \hline
    \textbf{Methods} & \textbf{mIoU} & \rotatebox{90}{reflection} &  \rotatebox{90}{vegetation} & \rotatebox{90}{ground} & \rotatebox{90}{road} & \rotatebox{90}{lane line} & \rotatebox{90}{stop Line} & \rotatebox{90}{sidewalk} & \rotatebox{90}{car} & \rotatebox{90}{truck}  &
    \rotatebox{90}{motorcycle} & \rotatebox{90}{pedestrian} & \rotatebox{90}{signs} & \rotatebox{90}{building}\\
    \hline
    \hline
    Cylinder3D \cite{zhu2021cylindrical} & 55.0 & 21.0 & 88.2 & 65.3 & 94.9 & 42.1 & 14.4 & 59.0 & 90.3 & 50.7 & 0.0 & 43.5 & 57.8 & 87.7 \\
    \hline
    SphereFormer \cite{lai2023spherical} & 63.5 & 23.4 & 92.5 & 77.2 & 97.3 & 57.0 & 33.4 & 73.9 & 93.3 & 56.5 & 0.0 & 59.9 & 69.6 & 91.9 \\
    \hline
    \hline
    Ours & \textbf{64.0} & 23.6 & 92.1 & 77.9 & 97.6 & 58.8 & 41.7 & 75.1 & 94.1 & 63.4 & 0.0 & 55.4 & 61.5 & 91.3 \\
    \hline
    \end{tabular}
    }
\end{table*}
\subsubsection{Hybrid-Solid LiDAR.}
 In \cref{tab:valall} and \cref{tab:S.MID}, we compare the results of our proposed method with open source SOTA LiDAR segmentation methods on our dataset S.MID. The point cloud distribution of each frame in hybrid-solid LiDAR exhibits \textbf{significant randomness} compared to mechanical spinning LiDAR. Previous methods demonstrate only marginal improvement compared to the common baseline SSCN \cite{graham20183d} (cubic and radial window attention with +0.2\% mIoU and cylindrical partition with +1.2\% mIoU), due to the failure of specially designed inductive bias. However, our method still outperforms 4.3\% mIoU. This demonstrates that when the distribution pattern of point clouds changes, our method is much more effective than previous approaches. 
 
\begin{table*}[t]
\caption{Results of our proposed method and SOTA LiDAR Segmentation methods on S.MID val set.  Note that all results are obtained from open source code with carefully chosen parameters.}
\label{tab:S.MID}
\centering
\resizebox{0.8\linewidth}{!}{
\begin{tabular}{c|c|c|c|c|c|c|c|c|c|c|c|c|c|c|c}
\hline
\textbf{Methods} & \textbf{mIoU} & \rotatebox{90}{knife switch} &  \rotatebox{90}{main xfmr} & \rotatebox{90}{arrester} & \rotatebox{90}{voltage xfmr} & \rotatebox{90}{busbar} & \rotatebox{90}{switch} & \rotatebox{90}{current xfmr} & \rotatebox{90}{scaffold} & \rotatebox{90}{sup column} & \rotatebox{90}{road} & \rotatebox{90}{other-ground} & \rotatebox{90}{fence} &
\rotatebox{90}{fire shelter} & \rotatebox{90}{wall} \\
\hline
\hline
SSCN \cite{graham20183d} (baseline) & 67.6 & 83.4 & 77.9 & 76.9 & 46.9 & 80.7 & 94.0 & 96,1 & 86.8 & 25.3 & 58.3 & 35.3 & 80.4 & 18.0 & 86.3 \\
\hline
Cylinder3D \cite{zhu2021cylindrical} & 68.8 (+1.2) & 83.7 & 77.8 & 74.8 & 41.5 & 79.6 & 93.5 & 94.2 & 83.6 & 48.9 & 54.5 & 35.8 & 81.2 & 29.1 & 84.7 \\
\hline
SphereFormer \cite{lai2023spherical} & 67.8 (+0.2) & 84.2 & 68.7 & 76.8 & 45.1 & 80.7 & 96.2 & 96.5 & 86.7 & 39.0 & 61.8 & 38.8 & 74.5 & 18.2 & 82.4 \\
\hline
\hline
Ours & \textbf{71.9 (+4.3)} & 89.1 & 91.9 & 85.0 & 43.0 & 77.6 & 96.6 & 96.3 & 88.2 & 52.4 & 60.3 & 33.4 & 85.1 & 19.9 & 88.7 \\
\hline
\end{tabular}
}
\end{table*}
\subsubsection{Visual Comparison.}
\begin{figure}[tb]
  \centering
  \includegraphics[width=\linewidth]{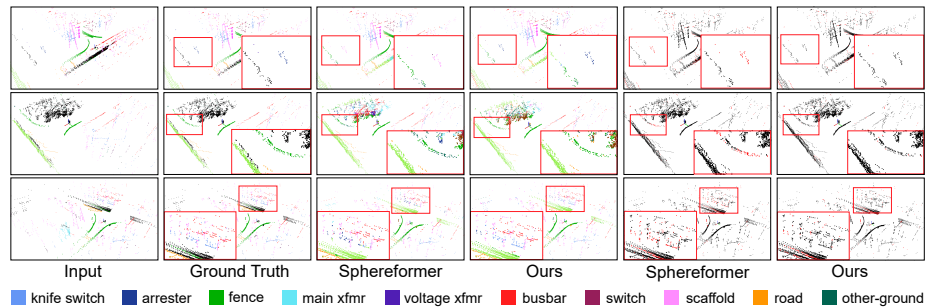}
  \caption{Visual comparison between SphereFormer \cite{lai2023spherical} and ours on S.MID val set. Details have been zoomed with red box. Difference maps are shown in the last two columns. More examples are given in the supplementary materials.}
  \label{fig:vis}
\end{figure}

In \cref{fig:vis}, we visually compare the results from SphereFormer \cite{lai2023spherical} and ours on S.MID. It visually indicates that our approach demonstrates superior performance in segmenting objects with similar geometric structures and distinguishing adjacent object boundaries.

\subsection{Network Analysis} \label{exp ana}
\subsubsection{Patterns for Multi-level Context Extraction.}
\begin{figure}[tb]
  \centering
  \includegraphics[width=\linewidth]{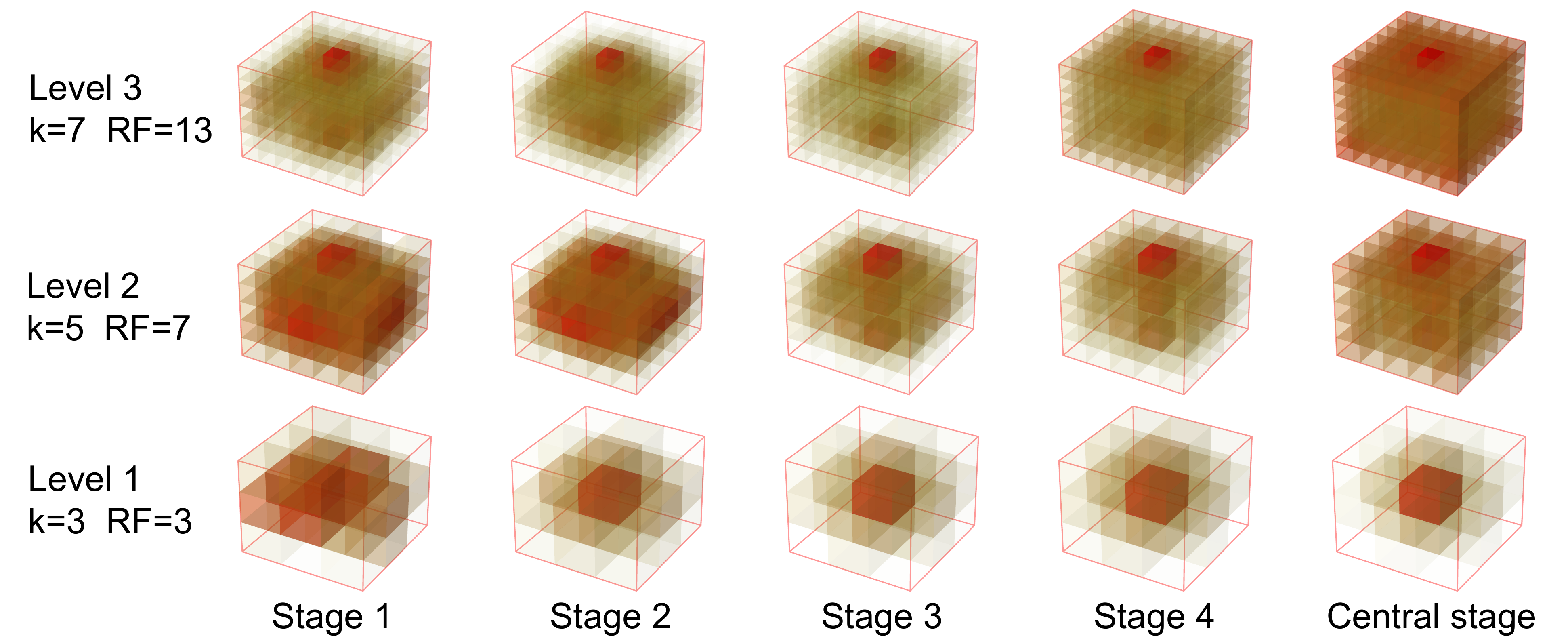}
  \caption{Visualization of parameters of $SubMconv^{l}_{3d}$ in \cref{eq:context} at three focal levels in four down stages and central stage in SFPNet on nuScenes. For general pattern display, we take the average of all channels. The small cubes in the image are depicted as being redder and more opaque, representing higher weights.}
  \label{fig:param}
\end{figure}

\cref{fig:param} shows the learned $SubMconv^{l}_{3d}$ kernels in our SFPNet on nuScenes. On the one hand, it demonstrates the adaptive feature capturing ability of our network. The models prioritize low focal levels to capture local features during early stages. As the stages progress, greater reliance is placed on long-range contexts. On the other hand, it shows the ability to accommodate point distributions. The horizontal resolution of mechanical spinning LiDAR is much higher than the vertical resolution. Therefore, the distributions in vertical direction and BEV are not equivalent in the representation space. The changes in weights also coincide with this property.

\subsubsection{Interpretability.}
\begin{figure}[tb]
  \centering
  \includegraphics[height=3.1cm]{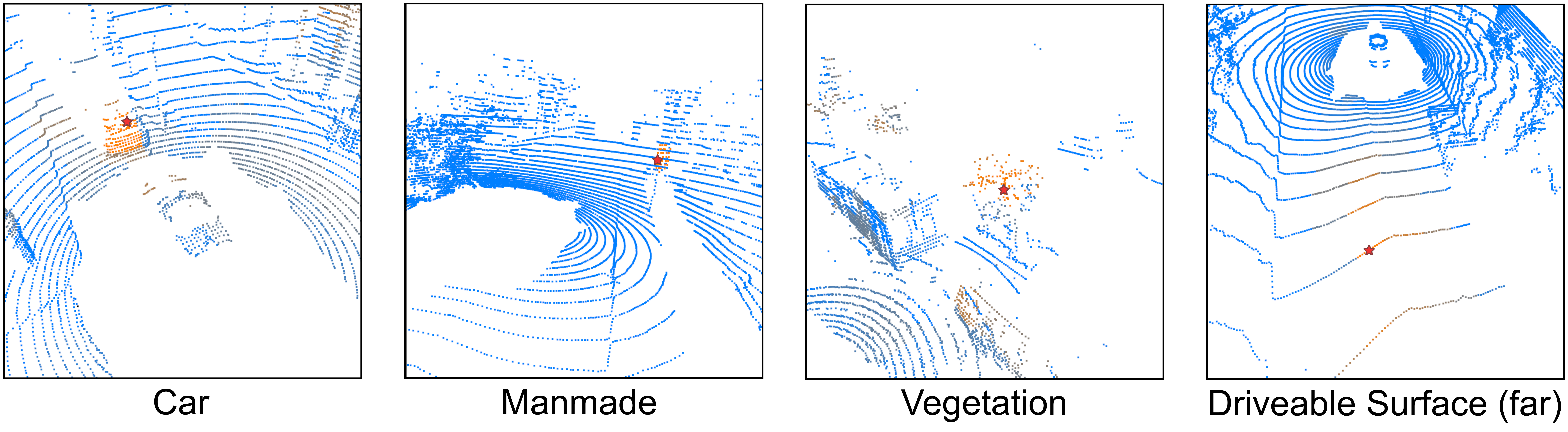}
  \caption{Visualization of correlations between a certain location and feature sequence in \cref{eq: int} at central stage on nuScenes val set. The results are restored through the correspondence with the input point clouds. Location $i$ has been marked as red star.}
  \label{fig:attn}
\end{figure}

\cref{fig:attn} illustrates the correlation between location $i$ and the feature sequence $X$ in high-dimensional space obtained from \cref{eq: int}. The visualization results simply elucidate the interpretability of our network. It is also noteworthy that even without utilizing a radial window or cylindrical partition, our model enables distant points to attend to a broader effective neighborhood.

\subsection{Ablation Study} \label{exp abl}
In order to evaluate the performance of each element within our networks, we carry out a series of ablation experiments utilizing the nuScenes dataset as shown in \cref{tab:Ablation Study}. We take SSCN \cite{graham20183d} as our basic blocks. By removing SFPM, a decrease of 4.7\% mIoU over the optimal design has been observed, which underscores the efficacy of SFPM.
\subsubsection{Fewer Focal Levels.}
We remove the last focal layer in SFPM and it hurts 1.9\% mIoU performance. This indicates that during multi-level context extraction, longer-range context can offer a larger effective receptive field, thereby compensating for the deficiency of local features faced by hard tokens due to sparsity.
\subsubsection{No Global Average Pooling.}
Removing the global average pooling results in a 2.0\% mIoU disparity compared to the optimal design. This demonstrates the significance of global information despite the presence of long-range context.

\begin{table}[tb]
  \caption{Ablation study.}
  \label{tab:Ablation Study}
  \centering
  \scalebox{0.75}{  
    \begin{tabular}{@{}l|l|l|l|l|l@{}}
        \hline
         & Basic blocks & Focal level = 2 & Focal level = 3 & Global Avg Pooling & mIoU \\
        \hline
         Optimal design & $\checkmark$ &  & $\checkmark$ & $\checkmark$ & 80.1 \\
        \hline
        \hline
         Ablation 1 & $\checkmark$ & $\checkmark$ &  & $\checkmark$ & 78.2 (-1.9) \\
         Ablation 2 & $\checkmark$ &  & $\checkmark$ &  & 78.1 (-2.0) \\
         Ablation 3 & $\checkmark$ &  &  &  &  75.4 (-4.7) \\
      \hline
    \end{tabular}
  }
\end{table}
\section{Conclusion}
Our work proposes a generalized framework SFPNet to accommodate various types of LiDAR prevalent in the market. Our approach integrates multi-level context extraction and a gate mechanism to effectively aggregate both local and global features. Furthermore, we employ a lightweight element-wise multiplication operation with query to ensure the preservation of channel-wise information. A large-scale hybrid-solid LiDAR segmentation dataset for industrial robot applications has been collected and annotated under professional guidance. Our proposed network demonstrates outstanding performance on datasets derived from various types of LiDAR and exhibits excellent interpretability. We anticipate a diversification of LiDAR technologies for future industrial applications. We believe that our proposed network can better adapt to various types of LiDAR sensors or even their fused data. The limitations of our work are discussed in the supplementary materials.

%
%

\bibliographystyle{splncs04}
\bibliography{main}

\clearpage
\title{Supplementary Materials for SFPNet}
\titlerunning{Supplementary Materials for SFPNet}

\author{} 

\authorrunning{Y.~Wang et al.}

\institute{}
\maketitle

\section{Introduction} \label{sec:intro}
In this supplementary materials, we provide our \textbf{dataset details} about sensors, scenes, annotation process and label distributions in \cref{dataset}.  Additional method details are demonstrated in \cref{meth}. More experiment results and network analysis are given in \cref{exp}. Limitations and future works are discussed in \cref{limit}.

\section{Dataset: \textbf{S}e\textbf{M}antic \textbf{I}n\textbf{D}ustry} \label{dataset}
\begin{table*}[t]
    \caption{Semantic LiDAR dataset comparison. Frames\textsuperscript{\dag} for train/val/test. Number of classes \textsuperscript{\ddag} for single frame evaluation and annotated total number in brackets.}
    \label{tab:comparedata}
    \centering
    \resizebox{\linewidth}{!}{
    \begin{tabular}{c|c|c|c|c|c}
    \hline
    \textbf{Datasets} & \textbf{Frames\textsuperscript{\dag}} & \textbf{LiDAR} & \textbf{Types of LiDAR} & \textbf{Classes\textsuperscript{\ddag}} & \textbf{Applications} \\
    \hline
    \hline
    nuScenes & 28130/6019/6008 & Velodyne-HDL-32E & Mechanical Spinning LiDAR & 16 (32) & Autonomous Vehicle \\
    \hline
    SemanticKITTI & 19130/4071/20351  & Velodyne-HDL-64E & Mechanical Spinning LiDAR & 19 (34) & Autonomous Vehicle\\
    \hline
    \hline
    S.MID & 13101/5000/20803 & Livox Mid-360 & \textbf{Hybrid-Solid LiDAR} & 14 (25)  & \textbf{Industrial Robot} \\
    \hline
    \end{tabular}
    }
\end{table*}
\subsection{Scenes}
\label{1}
Many applications rely on the crucial aspect of comprehending semantic scenes. However, most existing benchmarks \cite{behley2019semantickitti,caesar2020nuscenes,sun2020scalability,xiao2021pandaset} focus on driving scenes. To fill the gap in public dataset of industrial outdoor scenes for \textbf{robotic application}, we collect a total of 38904 frames of hybrid-solid LiDAR data in different substations and have annotated 25 categories as shown in \cref{fig:map}. Overall comparison with previous benchmarks is shown in \cref{tab:comparedata}.

\subsection{Sensors}
\begin{figure}[tb]
\centering
\includegraphics[width=\textwidth]{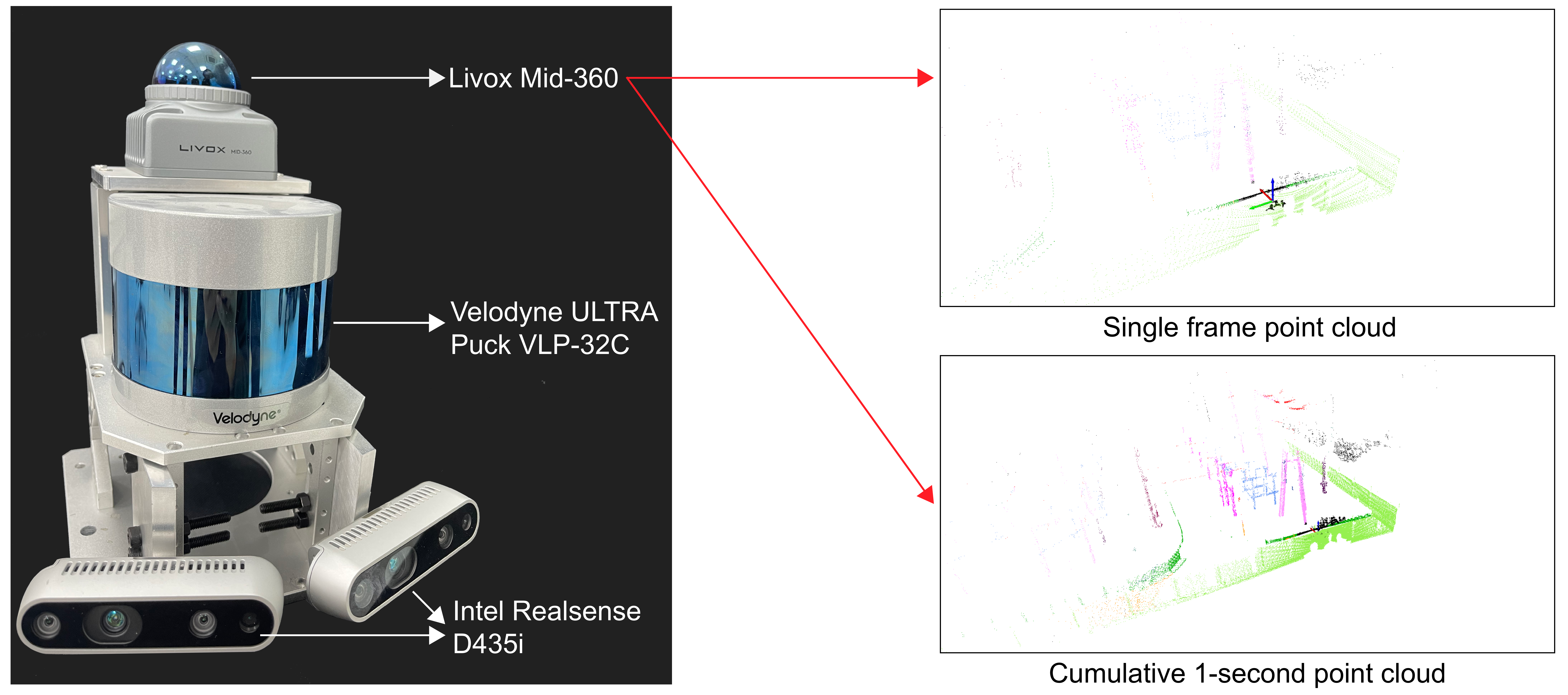}
\caption{Sensors and comparison between single frame and cumulative 1-second point clouds for Livox Mid-360. Although the single-frame point cloud is relatively sparse, the cumulative point cloud can better express the scene in the vertical direction. Please also note that only data collected by Livox Mid-360 and the corresponding labels are used in this research and have been released with S.MID.}
\label{fig:sensor}
\end{figure}

\cref{fig:sensor} shows the sensors equipped on our industrial robot used to collect S.MID. To the best of our knowledge, S.MID is the first large-scale outdoor \textbf{hybrid-solid LiDAR semantic segmentation dataset}. In addition to the features shown in the figures, Livox Mid-360 is much more cost-effective compared to traditional mechanical spinning LiDAR. 

In accordance with the illustration provided in \cref{fig:sensor} and Fig. 1 (b) in the main text, Livox Mid-360 is suitable for industrial robots involving scene understanding tasks since it covers a broader range of scenes with \textbf{non-repetitive scanning mode}. However, it is a double-edged sword. This mode will also make the point cloud relatively \textbf{sparse} and \textcolor{red}{\textbf{randomly distributed}}. Therefore, the single-frame hybrid-solid LiDAR segmentation task \textbf{brings more challenges to network design}. 

\subsection{Annotation Process}
\begin{figure}[tb]
\centering
\includegraphics[width=0.95\linewidth]{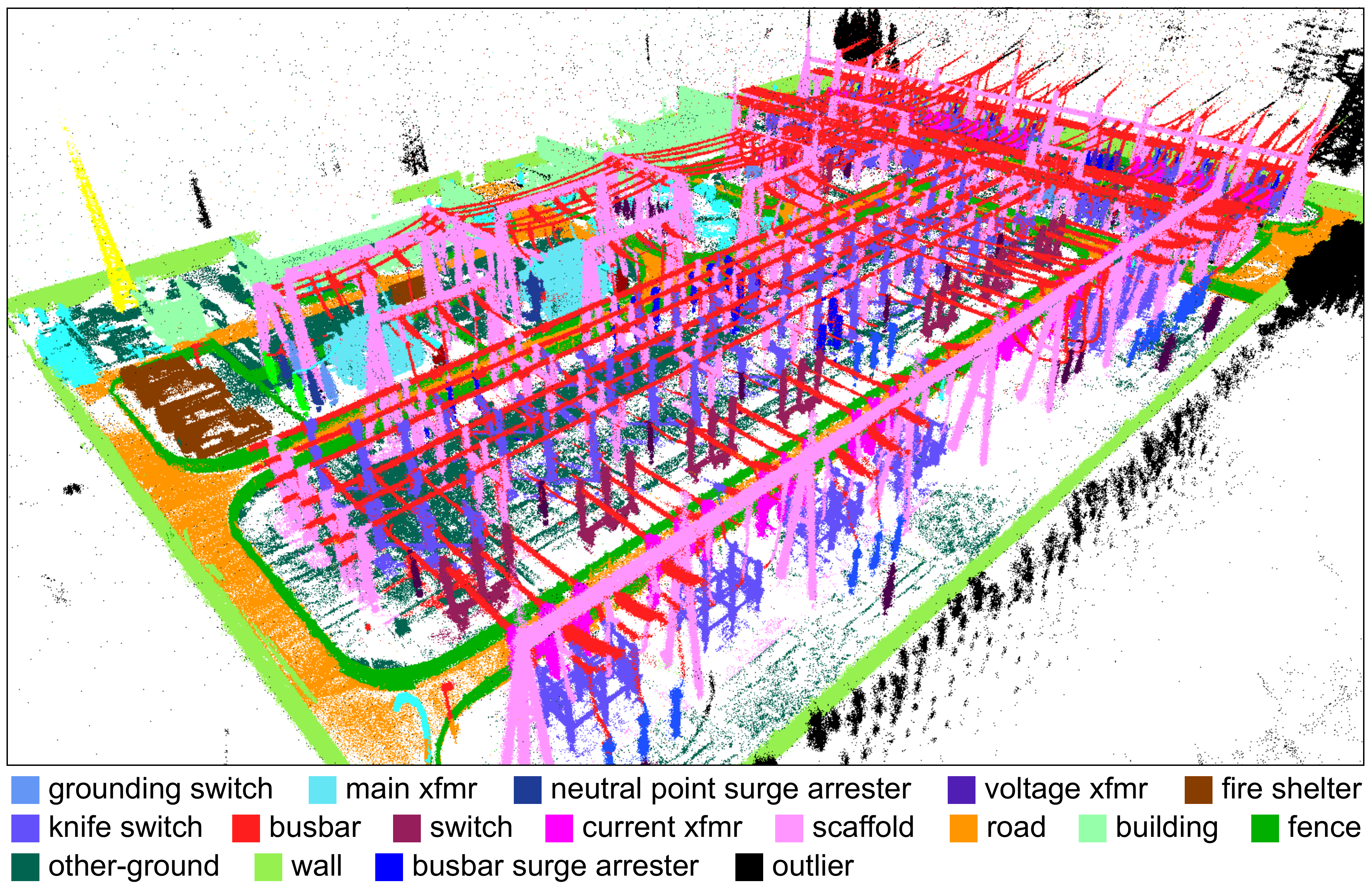}
\caption{Example of maps built in the annotation process.}
\label{fig:map}
\end{figure}

Considering the safety inspection tasks of robots and the common objects found in substations, we have annotated a total of 25 categories under professional guidance. Acknowledging the tools and annotation strategies provided by previous researchers \cite{behley2019semantickitti}, we first develop a high-precision LiDAR-inertial SLAM system based on hybrid-state LiDAR for initial mapping. 
Subsequently, through manual correction, high-precision maps for annotation purposes are obtained as shown in \cref{fig:map}.

Due to the presence of specialized equipment within the substations, there is a requirement for the annotators' expertise compared to that of annotators for autonomous driving datasets. Following training conducted by professionals, our dataset's labels have been carefully annotated.

\subsection{Label Distributions}
\begin{figure}[tb]
\centering
\includegraphics[width=0.95\textwidth]{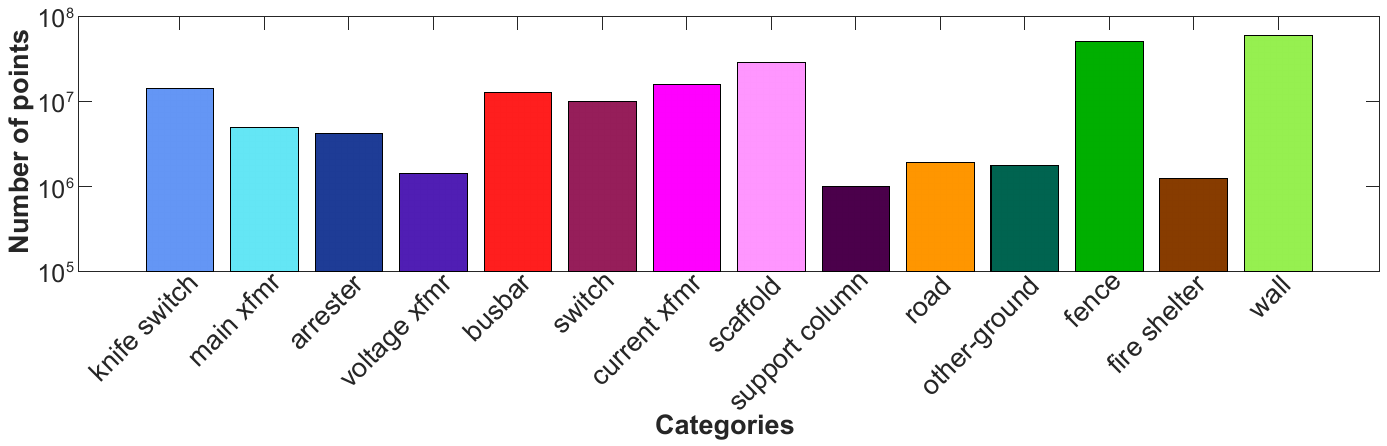}
\caption{Label distributions.}
\label{fig:label}
\end{figure}

For single-frame segmentation task, we merge the annotated labels into 14 classes (\textit{knife switch, main transformer, arrester, voltage transformer, busbar, switch, current transformer, scaffold, support column, road, other-ground, fence, fire shelter, wall}). The label distributions are shown in \cref{fig:label}. The imbalanced count of classes is common in substation scenes. 
Hence, similar to imbalanced class distributions observed in autonomous driving datasets, addressing the issue of imbalanced class distribution in S.MID is an integral aspect that methods must contend with.
\label{3}

\section{Additional Method Details} \label{meth}

\subsection{Overall Framework} \label{Framework Overview}
\begin{figure}[tb]
  \centering
  \includegraphics[width=\linewidth]{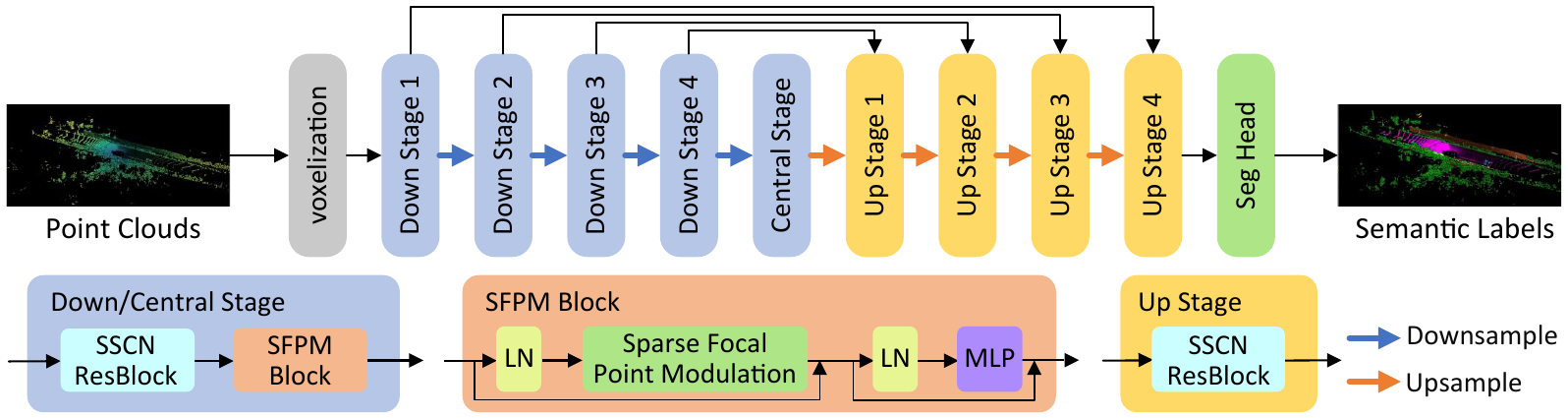}
  \caption{Overall Framework. Our network employs an encoder-decoder structure with four down/up stages and one central stage. Similar to the transformer \cite{vaswani2017attention}, our sparse focal point block consists of core modulator SFPM, layer normalization, and MLP as feed-forward network.}
  \label{fig:framework}
\end{figure}

Following the previous work \cite{zhu2021cylindrical,lai2023spherical}, we adopt a U-Net\cite{ronneberger2015u} structure as shown in \cref{fig:framework}. We firstly apply regular voxelization to form a sparse tensor $X \in \mathbb{R}^{N \times C_{in}}$. Our sparse focal point module is introduced in down stages and central stage. After traversing through the backbone with skip connections, we employ a simple projection head to get the segmentation result. Due to the long-tailed data distribution in the prevalent LiDAR semantic segmentation datasets, we adopt focal loss \cite{lin2017focal} to address the issue of class imbalance. 
\subsection{Properties Discussion} 
Proof of \textit{translation invariance} can be found in Sec. 3.1 in the main text. Here, we provide an extension analysis of \textit{explicit locality with contextual learning}. The realization of our aggregation step $\kappa_{focal} ( \cdot )$ is achieved through linear projection and Eqs. (4) – (6). The set of increasing kernels of SubMconv layers in Eq. (4) provides explicit locality and the operations before and after it will preserve this property (element-wise multiplication or channel-wise calculation). By using the gate mechanism described in Eq. (5), the input-dependent multi-level context from Eq. (4) can be adaptively aggregated. Additionally, Eq. (5) provides a ``soft shape'' in the sparse space through corresponding gate weight for each position $i$. Heuristic thinking: When dealing with diverse point cloud distributions, varying densities in each scan, and distinct classes, a qualified feature encoder exhibits varying dependencies across different contextual levels and positions within sparse space.

\section{Additional experiments} \label{exp}
\label{4}
\begin{table*}[t]
\caption{Results of our proposed method and SOTA LiDAR Segmentation methods on S.MID test set.  Note that all results are obtained from open source code with carefully chosen parameters.}
\label{tab:S.MID}
\centering
\resizebox{0.8\linewidth}{!}{
\begin{tabular}{c|c|c|c|c|c|c|c|c|c|c|c|c|c|c|c}
\hline
\textbf{Methods} & \textbf{mIoU} & \rotatebox{90}{knife switch} &  \rotatebox{90}{main xfmr} & \rotatebox{90}{arrester} & \rotatebox{90}{voltage xfmr} & \rotatebox{90}{busbar} & \rotatebox{90}{switch} & \rotatebox{90}{current xfmr} & \rotatebox{90}{scaffold} & \rotatebox{90}{sup column} & \rotatebox{90}{road} & \rotatebox{90}{other-ground} & \rotatebox{90}{fence} &
\rotatebox{90}{fire shelter} & \rotatebox{90}{wall} \\
\hline
\hline
Cylinder3D \cite{zhu2021cylindrical} & 68.1 & 82.9 & 69.8 & 74.8 & 44.1 & 79.1 & 92.9 & 93.5 & 79.9 & 54.7 & 57.0 & 37.9 & 77.6 & 28.4 & 81.0 \\
\hline
SphereFormer \cite{lai2023spherical} & 68.3 & 84.2 & 71.5 & 75.5 & 49.8 & 80.1 & 96.6 & 96.7 & 86.6 & 47.5 & 60.8 & 40.1 & 74.7 & 8.9 & 83.4 \\
\hline
\hline
Ours & \textbf{70.9} & 88.8 & 90.4 & 85.2 & 50.4 & 76.1 & 97.1 & 96.9 & 89.2 & 60.2 & 57.6 & 29.7 & 83.1 & 1.2 & 87.3 \\
\hline
\end{tabular}
}
\end{table*}

\begin{figure}[tb]
  \centering
  \includegraphics[width=\linewidth]{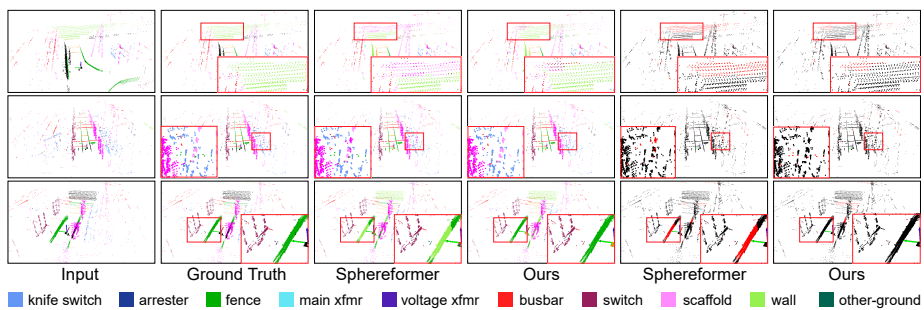}
  \caption{Visual comparison between SphereFormer \cite{lai2023spherical} and ours on S.MID val set. Details have been zoomed with red box. Difference maps are shown in the last two columns.}
  \label{visual}
\end{figure}

\begin{figure}[tb]
  \centering
  \subfloat[SemanticKITTI.]{\includegraphics[width=\linewidth]{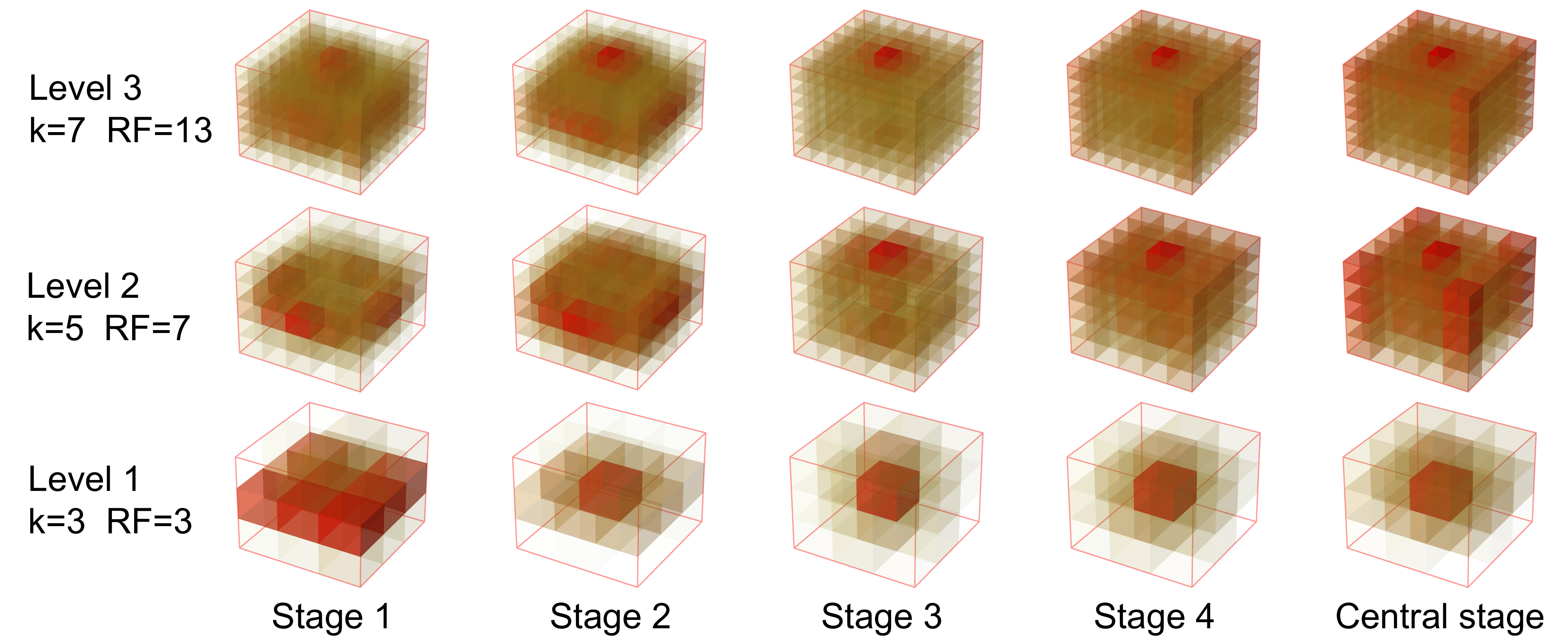}}
  \quad
  \subfloat[S.MID.]{\includegraphics[width=\linewidth]{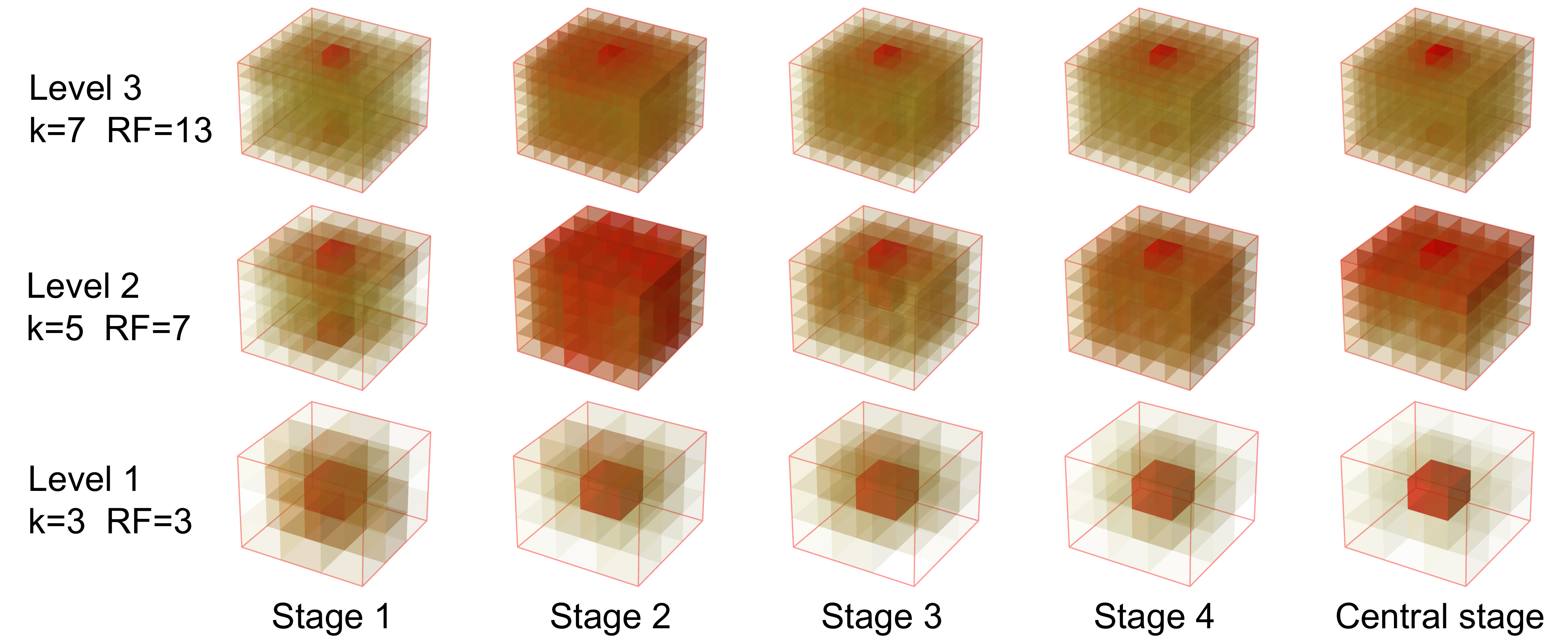}}
  \caption{Visualization of parameters of $SubMconv^{l}_{3d}$ at three focal levels in four down stages and central stage in SFPNet. SemanticKITTI shows similar patterns to nuScenes as demonstrate in the main text. S.MID shows a special pattern in the vertical direction due to the particularity of its point cloud.}
  \label{param}
\end{figure}

More segmentation results on SemanticKITTI val and test sets are displayed in \cref{kitval,kittest} and nuScenes test set in \cref{nustest} and S.MID test set in \cref{tab:S.MID} Additional ablation study on S.MID in \cref{tab:Ablation Study2}. More visual comparisons between SphereFormer \cite{lai2023spherical} and ours on S.MID val set are shown in \cref{visual}. More network analysis results are shown in \cref{param}.

Since most of the previous training techniques and augmentation methods such as Cutmix \cite{xu2021rpvnet,lai2023spherical}, Lasermix \cite{kong2023lasermix}, Polarmix \cite{xiao2022polarmix} and post-processing \cite{kong2023rethinking} are designed for mechanical spinning LiDAR, in order to ensure the consistency of the three different types of LiDAR experiments, we did not use any training techniques. In this situation, SFPNet still shows competitive results on mechanical spinning LiDAR test sets. 

In both S.MID val (in the main text) and test set (\cref{tab:S.MID}), we can see that when the distribution pattern of point clouds changes, the performance of cubic and radial window attention will deteriorate or even become worse than that of the improved SSCN. This shows that SFPM can better cope with different types of LiDAR with various point distributions due to its adaptive mechanism.

\begin{table*}[t]
    \caption{Results of our proposed method and state-of-the-art LiDAR Segmentation methods on SemanticKITTI val set. Note that all results are obtained from the literature.}
    \label{kitval}
    \centering
    \resizebox{0.75\linewidth}{!}{
    \begin{tabular}{c|c|c|c|c|c|c|c|c|c|c|c|c|c|c|c|c|c|c|c|c}
    \hline
    \textbf{Methods} & \textbf{mIoU} & \rotatebox{90}{car} &  \rotatebox{90}{bicycle} & \rotatebox{90}{motor.} & \rotatebox{90}{truck} & \rotatebox{90}{other-veh.} & \rotatebox{90}{person} & \rotatebox{90}{bicyclist} & \rotatebox{90}{m.cyclist} & \rotatebox{90}{road} & \rotatebox{90}{parking} & \rotatebox{90}{sidewalk} & \rotatebox{90}{other-gro.} &
    \rotatebox{90}{building} & \rotatebox{90}{fence} & \rotatebox{90}{vegetation} & \rotatebox{90}{trunk} & \rotatebox{90}{terrain} & \rotatebox{90}{pole} & \rotatebox{90}{traffic s.} \\
    \hline
    \hline
    SSCN \cite{graham20183d} & 66.6 & 96.3 & 44.6 & 76.3 & 89.6 & 58.6 & 77.3 & 91.3 & 0.0 & 94.3 & 51.7 & 81.8 & 1.2 & 91.0 & 62.5 & 88.3 & 70.2 & 75.3 & 64.6 & 51.4
    \\ \hline
    SphereFormer \cite{lai2023spherical} & 69.0 & 97.0 & 53.4 & 77.2 & 95.1 & 67.0 & 78.2 & 93.7 & 0.0 & 95.2 & 55.5 & 83.1 & 2.8 & 91.0 & 60.4 & 89.2 & 72.5 & 76.9 & 66.3 & 55.9 \\
    \hline
    \hline
    Ours & \textbf{69.2} & 97.2 & 53.2 & 80.2 & 93.1 & 70.6 & 75.4 & 91.5 & 0.0 & 95.2 & 56.3 & 83.4 & 3.3 & 92.2 & 66.8 & 89.3 & 72.6 & 76.7 & 65.0 & 51.9 \\
    \hline
    \end{tabular}
    }
\end{table*}

\section{Limitations and Future works} \label{limit}
Our work focuses on the representational capabilities of the network on general LiDAR point clouds. However, data augmentation, training techniques and post-processing are also important topics for segmentation tasks. For instance, 3.7\% \~{} 4.9\% mIoU improvement for SSCN-based networks can be achieved on mechanical spinning LiDAR through Polarmix \cite{xiao2022polarmix} . 

Future works can be done to explore augmentation methods for general LiDAR point clouds. We will also extend our methods to more LiDAR point cloud tasks such as object detection and panoptic segmentation, and on fused various types of LiDAR datasets. Efficiency improvement will also be considered in the future. \label{6}


\clearpage  

\begin{table*}[t]
    \caption{Results of our proposed method and state-of-the-art LiDAR Segmentation methods on SemanticKITTI test set. Note that all results are obtained from the literature. LiDAR-based methods in the table are listed by year of publication.}
    \label{kittest}
    \centering
    \resizebox{0.75\linewidth}{!}{
    \begin{tabular}{c|c|c|c|c|c|c|c|c|c|c|c|c|c|c|c|c|c|c|c|c}
    \hline
    \textbf{Methods} & \textbf{mIoU} & \rotatebox{90}{car} &  \rotatebox{90}{bicycle} & \rotatebox{90}{motor.} & \rotatebox{90}{truck} & \rotatebox{90}{other-veh.} & \rotatebox{90}{person} & \rotatebox{90}{bicyclist} & \rotatebox{90}{m.cyclist} & \rotatebox{90}{road} & \rotatebox{90}{parking} & \rotatebox{90}{sidewalk} & \rotatebox{90}{other-gro.} &
    \rotatebox{90}{building} & \rotatebox{90}{fence} & \rotatebox{90}{vegetation} & \rotatebox{90}{trunk} & \rotatebox{90}{terrain} & \rotatebox{90}{pole} & \rotatebox{90}{traffic s.} \\
    \hline
    \hline
    PointNet++ \cite{qi2017pointnet++} & 20.1 & 53.7 & 1.9 & 0.2 & 0.9 & 0.2 & 0.9 & 1.0 & 0.0 & 72.0 & 18.7 & 41.8 & 5.6 & 62.3 & 16.9 & 46.5 & 13.8 & 30.0 & 6.0 & 8.9
    \\ \hline
    TangentConv \cite{tatarchenko2018tangent} & 40.9 & 90.8 & 2.7 & 16.5 & 15.2 & 12.1 & 23.0 & 28.4 & 8.1 & 83.9 & 33.4 & 63.9 & 15.4 & 83.4 & 49.0 & 79.5 & 49.3 & 58.1 & 35.8 & 28.5
    \\ \hline
    SqueezeSegV2 \cite{wu2019squeezesegv2} & 39.7 & 81.8 & 18.5 & 17.9 & 13.4 & 14.0 & 20.1 & 25.1 & 3.9 & 88.6 & 45.8 & 67.6 & 17.7 & 73.7 & 41.1 & 71.8 & 35.8 & 60.2 & 20.2 & 26.3
    \\ \hline
    DarkNet53Seg \cite{behley2019semantickitti} & 49.9 & 86.4 & 24.5 & 32.7 & 25.5 & 22.6 & 36.2 & 33.6 & 4.7 & 91.8 & 64.8 & 74.6 & 27.9 & 84.1 & 55.0 & 78.3 & 50.1 & 64.0 & 38.9 & 52.2
    \\ \hline
    RangeNet53++ \cite{milioto2019rangenet++} & 52.2 & 91.4 & 25.7 & 34.4 & 25.7 & 23.0 & 38.3 & 38.8 & 4.8 & 91.8 & 65.0 & 75.2 & 27.8 & 87.4 & 58.6 & 80.5 & 55.1 & 64.6 & 47.9 & 55.9
    \\ \hline
    KPConv \cite{thomas2019kpconv} & 58.8 & 95.0 & 30.2 & 42.5 & 33.4 & 44.3 & 61.5 & 61.6 & 11.8 & 90.3 & 61.3 & 72.7 & 31.5 & 90.5 & 64.2 & 84.8 & 69.2 & 69.1 & 56.4 & 47.4
    \\ \hline
    3D-MiniNet \cite{alonso20203dmininet} & 55.8 & 90.5 & 42.3 & 42.1 & 28.5 & 29.4 & 47.8 & 44.1 & 14.5 & 91.6 & 64.2 & 74.5 & 25.4 & 89.4 & 60.8 & 82.8 & 60.8 & 66.7 & 48.0 & 56.6
    \\ \hline
    SqueezeSegV3 \cite{xu2020squeezesegv3} & 55.9 & 92.5 & 38.7 & 36.5 & 29.6 & 33.0 & 45.6 & 46.2 & 20.1 & 91.7 & 63.4 & 74.8 & 26.4 & 89.0 & 59.4 & 82.0 & 58.7 & 65.4 & 49.6 & 58.9
    \\ \hline
    PointASNL \cite{yan2020pointasnl} & 46.8 & 87.9 & 0.0 & 25.1 & 39.0 & 29.2 & 34.2 & 57.6 & 0.0 & 87.4 & 24.3 & 74.3 & 1.8 & 83.1 & 43.9 & 84.1 & 52.2 & 70.6 & 57.8 & 36.9
    \\ \hline
    RandLA-Net \cite{hu2020randla} & 55.9 & 94.2 & 29.8 & 32.2 & 43.9 & 39.1 & 48.4 & 47.4 & 9.4 & 90.5 & 61.8 & 74.0 & 24.5 & 89.7 & 60.4 & 83.8 & 63.6 & 68.6 & 51.0 & 50.7
    \\ \hline
    PolarNet \cite{zhang2020polarnet} & 54.3 & 93.8 & 40.3 & 30.1 & 22.9 & 28.5 & 43.2 & 40.2 & 5.6 & 90.8 & 61.7 & 74.4 & 21.7 & 90.0 & 61.3 & 84.0 & 65.5 & 67.8 & 51.8 & 57.5
    \\ \hline
    SPVNAS \cite{tang2020searching} & 67.0 & 97.2 & 50.6 & 50.4 & 56.6 & 58.0 & 67.4 & 67.1 & 50.3 & 90.2 & 67.6 & 75.4 & 21.8 & 91.6 & 66.9 & 86.1 & 73.4 & 71.0 & 64.3 & 67.3
    \\ \hline
    JS3C-Net \cite{yan2021sparse} & 66.0 & 95.8 & 59.3 & 52.9 & 54.3 & 46.0 & 69.5 & 65.4 & 39.9 & 88.9 & 61.9 & 72.1 & 31.9 & 92.5 & 70.8 & 84.5 & 69.8 & 67.9 & 60.7 & 68.7
    \\ \hline
    Cylinder3D \cite{zhu2021cylindrical} & 68.9 & 97.1 & 67.6 & 63.8 & 50.8 & 58.5 & 73.7 & 69.2 & 48.0 & 92.2 & 65.0 & 77.0 & 32.3 & 90.7 & 66.5 & 85.6 & 72.5 & 69.8 & 62.4 & 66.2
    \\ \hline
    (AF)$^{2}$-S3Net \cite{cheng20212} & 70.8 & 94.3 & 63.0 & 81.4 & 40.2 & 40.0 & 76.4 & 81.7 & 77.7 & 92.0 & 66.8 & 76.2 & 45.8 & 92.5 & 69.6 & 78.6 & 68.0 & 63.1 & 64.0 & 73.3
    \\ \hline
    RPVNet \cite{xu2021rpvnet} & 70.3 & 97.6 & 68.4 & 68.7 & 44.2 & 61.1 & 75.9 & 74.4 & 43.4 & 93.4 & 70.3 & 80.7 & 33.3 & 93.5 & 72.1 & 86.5 & 75.1 & 71.7 & 64.8 & 61.4
    \\ \hline
    RangeViT-CS \cite{ando2023rangevit} & 64.0 & 95.4 & 55.8 & 43.5 & 29.8 & 42.1 & 63.9 & 58.2 & 38.1 & 93.1 & 70.2 & 80.0 & 32.5 & 92.0 & 69.0 & 85.3 & 70.6 & 71.2 & 60.8 & 64.7
    \\ \hline
    RangeFormer\cite{kong2023rethinking} & 73.3 & 96.7 & 69.4 & 73.7 & 59.9 & 66.2 & 78.1 & 75.9 & 58.1 & 92.4 & 73.0 & 78.8 & 42.4 & 92.3 & 70.1 & 86.6 & 73.3 & 72.8 & 66.4 & 66.6
    \\ \hline
    SphereFormer \cite{lai2023spherical} & 74.8 & 97.5 & 70.1 & 70.5 & 59.6 & 67.7 & 79.0 & 80.4 & 75.3 & 91.8 & 69.7 & 78.2 & 41.3 & 93.8 & 72.8 & 86.7 & 75.1 & 72.4 & 66.8 & 72.9 \\
    \hline
    \hline
    Ours & 70.3 & 97.2 & 64.9 & 63.8 & 44.8 & 54.7 & 70.4 & 74.6 & 52.9 & 91.9 & 70.6 & 78.0 & 39.7 & 93.3 & 71.5 & 85.4 & 73.7 & 70.1 & 66.1 & 72.1 \\
    \hline
    \end{tabular}
    }
\end{table*}

\begin{table*}[t]
\caption{Results of our proposed method and state-of-the-art LiDAR Segmentation methods on nuScenes test set. Note that all results are obtained from the literature. Methods in the table are listed by year of publication.}
\label{nustest}
\centering
\resizebox{0.75\linewidth}{!}{
\begin{tabular}{c|c|c|c|c|c|c|c|c|c|c|c|c|c|c|c|c|c|c|c}
\hline
\textbf{Methods}  & \textbf{Input} &\textbf{mIoU} & \textbf{FW mIoU} & \rotatebox{90}{barrier} &  \rotatebox{90}{bicycle} & \rotatebox{90}{bus} & \rotatebox{90}{car} & \rotatebox{90}{construction} & \rotatebox{90}{motor} & \rotatebox{90}{pedestrian} & \rotatebox{90}{traffic cone} & \rotatebox{90}{trailer} & \rotatebox{90}{truck} & \rotatebox{90}{driveable} & \rotatebox{90}{other flat} &
\rotatebox{90}{sidewalk} & \rotatebox{90}{terrain} & \rotatebox{90}{manmade} & \rotatebox{90}{vegetation} \\
\hline
\hline
PolarNet \cite{zhang2020polarnet} & L & 69.4 & 87.4 & 72.2 & 16.8 & 77.0 & 86.5 & 51.1 & 69.7 & 64.8 & 54.1 & 69.7 & 63.5 & 96.6 & 67.1 & 77.7 & 72.1 & 78.1 & 84.5
\\ \hline
AMVNet \cite{Liong2020AMVNetAM} & L & 77.3 & 90.1 & 80.6 & 32.0 & 81.7 & 88.9 & 67.1 & 84.3 & 76.1 & 73.5 & 84.9 & 67.3 & 97.5 & 67.4 & 79.4 & 75.5 & 91.5 & 88.7
\\ \hline
SPVCNN \cite{tang2020searching} & L & 77.4 & 89.7 & 80.0 & 30.0 & 91.9 & 90.8 & 64.7 & 79.0 & 75.6 & 70.9 & 81.0 & 74.6 & 97.4 & 69.2 & 80.0 & 76.1 & 89.3 & 87.1
\\ \hline
JS3C-Net \cite{yan2021sparse} & L & 73.6 & 88.1 & 80.1 & 26.2 & 87.8 & 84.5 & 55.2 & 72.6 & 71.3 & 66.3 & 76.8 & 71.2 & 96.8 & 64.5 & 76.9 & 74.1 & 87.5 & 86.1
\\ \hline
Cylinder3D \cite{zhu2021cylindrical} & L & 77.2 & 89.9 & 82.8 & 29.8 & 84.3 & 89.4 & 63.0 & 79.3 & 77.2 & 73.4 & 84.6 & 69.1 & 97.7 & 70.2 & 80.3 & 75.5 & 90.4 & 87.6
\\ \hline
(AF)$^{2}$-S3Net \cite{cheng20212} & L & 78.3 & 88.5 & 78.9 & 52.2 & 89.9 & 84.2 & 77.4 & 74.3 & 77.3 & 72.0 & 83.9 & 73.8 & 97.1 & 66.5 & 77.5 & 74.0 & 87.7 & 86.8
\\ \hline
PMF \cite{zhuang2021perception} & L+C & 77.0 & 89.0 & 82.0 & 40.0 & 81.0 & 88.0 & 64.0 & 79.0 & 80.0 & 76.0 & 81.0 & 67.0 & 97.0 & 68.0 & 78.0 & 74.0 & 90.0 & 88.0
\\ \hline
2D3DNet \cite{genova20212d3dnet} & L+C & 80.0 & 90.1 & 83.0 & 59.4 & 88.0 & 85.1 & 63.7 & 84.4 & 82.0 & 76.0 & 84.8 & 71.9 & 96.9 & 67.4 & 79.8 & 76.0 & 92.1 & 89.2
\\ \hline
RangeFormer\cite{kong2023rethinking} & L & 80.1 & 90.0 & 85.6 & 47.4 & 91.2 & 90.9 & 70.7 & 84.7 & 77.1 & 74.1 & 83.2 & 72.6 & 97.5 & 70.7 & 79.2 & 75.4 & 91.3 & 88.9
\\ \hline
SphereFormer \cite{lai2023spherical} & L & 81.9 & 91.7 & 83.3 & 39.2 & 94.7 & 92.5 & 77.5 & 84.2 & 84.4 & 79.1 & 88.4 & 78.3 & 97.9 & 69.0 & 81.5 & 77.2 & 93.4 & 90.2
\\ \hline
\hline
Ours & L & 80.2 & 90.8 & 83.7 & 42.5 & 89.1 & 91.5 & 74.1 & 83.5 & 79.1 & 74.7 & 87.3 & 73.3 & 97.7 & 78.1 & 80.3 & 76.2 & 92.3 & 89.3\\
\hline
\end{tabular}
}
\end{table*}

\begin{table}[tb]
  \caption{Additional ablation study on S.MID val set.}
  \label{tab:Ablation Study2}
  \centering
  \scalebox{0.75}{  
    \begin{tabular}{@{}l|l|l|l|l|l@{}}
        \hline
         & Basic blocks & Focal level = 2 & Focal level = 3 & Global Avg Pooling & mIoU \\
        \hline
         Optimal design & $\checkmark$ &  & $\checkmark$ & $\checkmark$ & 71.9 \\
        \hline
        \hline
         Ablation 1 & $\checkmark$ & $\checkmark$ &  & $\checkmark$ & 69.9 (-2.0) \\
         Ablation 2 & $\checkmark$ &  & $\checkmark$ &  & 69.8 (-2.1) \\
         Ablation 3 & $\checkmark$ &  &  &  &  67.6 (-4.3) \\
      \hline
    \end{tabular}
  }
\end{table}

\clearpage

\label{5}


%
%
\end{document}